\documentclass[oneside,10pt]{article}

\usepackage{times,url,mathrsfs}
\usepackage{amsmath}
\usepackage{amsfonts}
\usepackage{graphicx}
\usepackage{subfigure}
\newtheorem{theorem}{Theorem}

\begin{document}
\title{Training Logistic Regression and  SVM on 200GB Data Using b-Bit Minwise Hashing and Comparisons with Vowpal Wabbit (VW)}

\author{ Ping Li \\
         Dept. of Statistical Science\\
       Cornell University\\
         Ithaca, NY 14853\\
       pingli@cornell.edu
       \and
       Anshumali Shrivastava\\
         Dept. of Computer Science\\
         Cornell University\\
         Ithaca, NY 14853\\
         anshu@cs.cornell.edu
         \and
         Arnd Christian K\"{o}nig\\
         Microsoft Research\\
        Microsoft Corporation\\
        Redmond, WA 98052\\
        chrisko@microsoft.com
        }

\date{}
\maketitle

\begin{abstract}

\noindent  Our recent work on large-scale learning using $b$-bit minwise hashing~\cite{Report:Li_Moore_Konig_HashingSVM,Report:HashLearning11} was tested on the {\em webspam} dataset (about 24 GB in LibSVM format), which may be way too small compared to real datasets used in industry. Since we could not access the proprietary dataset used in~\cite{Proc:Weinberger_ICML2009} for testing the Vowpal Wabbit (VW) hashing algorithm, in this paper we present an experimental study based on the expanded {\em rcv1} dataset (about 200 GB in LibSVM format).

In our earlier report~\cite{Report:HashLearning11}, the experiments demonstrated that, with merely 200 hashed values per data point, $b$-bit minwise hashing can achieve similar test accuracies as VW with $10^6$ hashed values per data point, on the {\em webspam} dataset. In this paper,  our new experiments on the (expanded) {\em rcv1} dataset clearly agree with our earlier observation that  $b$-bit minwise hashing algorithm is substantially more accurate than VW hashing algorithm at the same storage. For example, with $2^{14}$ (16384) hashed values per data point, VW achieves similar test accuracies as $b$-bit hashing with merely 30 hashed values per data point. This is of course not surprising as the report~\cite{Report:HashLearning11} has already demonstrated that the variance of the VW algorithm can be order of magnitude(s) larger than the variance of $b$-bit minwise hashing. It was shown in \cite{Report:HashLearning11} that VW has the same variance as random projections. \\

At least in the context of search, minwise hashing has been widely used in industry. It is well-understood that the preprocessing cost is not a major issue because the preprocessing is trivially parallelizable and can be conducted off-line or combined with the data-collection process.  Nevertheless, in this paper, we report that, even merely from the perspective of academic machine learning practice, the preprocessing cost is not a major issue for the following reasons:
\begin{itemize}
\item The preprocessing incurs only a one-time cost. The same processed data can be used for many training experiments, for example, for many different ``C'' values in SVM cross-validation, or for different combinations of data splitting (into training and testing sets).

\item For training truly large-scale datasets, the dominating cost is often the data loading time. In our 200 GB dataset (which may be still very small according to the industry standard), the preprocessing cost  of $b$-bit minwise hashing is  on the same order of magnitude as the data loading time.
\item Using a GPU,  the preprocessing cost can be easily     reduced to a small fraction (e.g.,$<1/7$) of the data loading time.
\end{itemize}

The standard industry practice of minwise hashing is to use universal hashing to replace permutations. In other words, there is no need to store any permutation mappings, one of the reasons why minwise hashing is popular.  In this paper, we also provide experiments to verify this practice, based on the simplest 2-universal hashing, and illustrate that  the performance of $b$-bit minwise hashing does not degrade.

\end{abstract}

\section{Introduction}

Many machine learning applications are faced with  large and inherently high-dimensional datasets. For example, \cite{GoogleBlog} discusses training datasets with (on average) $10^{11}$ items and $10^9$ distinct features.  \cite{Proc:Weinberger_ICML2009} experimented with a dataset of potentially 16 trillion ($1.6\times10^{13}$) unique features. Interestingly, while large-scale learning has become a very urgent, hot topic, it is usually very difficult for researchers from universities to obtain truly large, high-dimensional datasets from industry. For example, the experiments in our recent work~\cite{Report:Li_Moore_Konig_HashingSVM,Report:HashLearning11} on large-scale learning using $b$-bit minwise hashing~\cite{Proc:Li_Konig_WWW10,Proc:Li_Konig_NIPS10,Article:Li_Konig_CACM11} were based on the {\em webspam} dataset (about 24 GB in LibSVM format), which may be too small.

To overcome this difficulty, we have generated a dataset of about 200 GB (in LibSVM format) from the {\em rcv1} dataset, using the original features + all pairwise combinations of features + 1/30 of the 3-way combinations of features. We choose 200 GB (which of course is still very small) because relatively inexpensive workstations with 192 GB memory are in the market, which may make it possible for LIBLINEAR~\cite{Article:Fan_JMLR08,Proc:Hsieh_ICML08}, the popular solver for logistic regression and linear SVM, to perform the training of the entire dataset in main memory. We hope in the near future we will be able to purchase such a workstation. Of course, in this ``information explosion'' age, the growth of data is always much faster than the growth of memory capacity.

Note that the our hashing method is orthogonal to particular solvers of logistic regression and SVM. We have tested $b$-bit minwise hashing with other solvers~\cite{Proc:Joachims_KDD06,Proc:Shalev-Shwartz_ICML07,URL:Bottou_SGD} and observed substantial improvements. We choose LIBLINEAR~\cite{Article:Fan_JMLR08} as the work horse because it is a popular tool and may be familiar to non-experts. Our experiments may be easily validated by simply generating the hashed data off-line and feeding them to LIBLINEAR (or other solvers) without  modification to the code. Also, we notice that the source code of LIBLINEAR, unlike many other excellent solvers, can be compiled  in Visual Studio without  modification. As many practitioners are using WINDOWS~\footnote{Note that the current version of Cygwin has a very serious memory limitation and hence is not suitable for large-scale experiments, even though all popular solvers can be compiled under Cygwin.}, we use LIBLINEAR throughout the paper, for the sake of maximizing the repeatability of our work.\\

Unsurprisingly, our experimental results agree with our prior studies~\cite{Report:HashLearning11} that $b$-bit minwise hashing  is substantially more accurate than the Vowpal Wabbit (VW) hashing algorithm~\cite{Proc:Weinberger_ICML2009} at the same storage. Note that in our paper, VW refers to the particular hashing algorithm in~\cite{Proc:Weinberger_ICML2009}, not the online learning platform that the authors of~\cite{Proc:Weinberger_ICML2009,Article:Shi_JMLR09} have been developing. For evaluation purposes, we must separate out hashing algorithms from learning algorithms because they are orthogonal to each other. \\

All randomized algorithms including minwise hashing and random projections rely on pseudo-random numbers. A common practice of minwise hashing (e.g.,~\cite{Proc:Broder}) is to use universal hashing functions to replace perfect random permutations. In this paper, we also present an empirical study to verify that this common practice does not degrade  the learning performance.

Minwise hashing has been widely deployed in industry and $b$-bit minwise hashing requires only minimal modifications. It is well-understood at least in the context of search that the (one time) preprocessing cost is not a major issue because the preprocessing step, which is trivially parallelizable, can be conducted off-line or combined in the data-collection process. In the context of pure machine learning research, one thing we notice is that for training truly large-scale datasets, the data loading time is often dominating~\cite{Proc:Yu_KDD10}, for online algorithms as well as batch algorithms (if the data fit in memory). Thus, if we have to load the data many times, for example, for testing different ``$C$'' values in SVM or running an online algorithms for multiple epoches, then the benefits of data reduction algorithms such as $b$-bit minwise hashing would be enormous.

Even on our dataset of 200 GB only, we observe that the preprocessing cost is roughly on the same order of magnitude as the data loading time. Furthermore, using a GPU (which is inexpensive) for fast hashing, we can reduce the preprocessing cost of $b$-bit minwise hashing to a small fraction of the data loading time. In other words, the dominating cost is the still the data loading time. \\

We are currently experimenting $b$-bit minwise hashing for machine learning with $\gg$ TB datasets and the results will be reported in subsequent technical reports. It is a very fun process to experiment with $b$-bit minwise hashing and we certainly would like to share our experience with the machine learning and data mining community.

\section{Review Minwise Hashing and b-Bit Minwise Hashing}\label{sec_minwise}

{\em Minwise hashing}~\cite{Proc:Broder,Proc:Broder_WWW97} has been successfully applied to a very wide range of real-world problems especially in the context of search~\cite{Proc:Broder,Proc:Broder_WWW97,Proc:Bendersky_WSDM09,Article:Forman09,Proc:Cherkasova_KDD09,Proc:Buehrer_WSDM08,Article:Urvoy08,Proc:Nitin_WSDM08,
Article:Dourisboure09,Proc:Chierichetti_KDD09,Proc:Gollapudi_WWW09,Article:Kalpakis08,Proc:Najork_WSDM09}, for efficiently computing set similarities.

 Minwise hashing mainly works well with binary data, which can be viewed either as 0/1  vectors or as sets. Given two sets, $S_1, \  S_2 \subseteq \Omega = \{0, 1, 2, ..., D-1\}$, a widely used (normalized)  measure of similarity is the {\em resemblance} $R$:
\begin{align}\notag
&R = \frac{|S_1 \cap S_2|}{|S_1 \cup S_2|} = \frac{a}{f_1 + f_2 - a}, \hspace{0.3in}
\text{where}\ \  f_1 = |S_1|, \  f_2 = |S_2|, \ a = |S_1\cap S_2|.
\end{align}
In this method, one applies a random permutation $\pi: \Omega\rightarrow\Omega$ on $S_1$ and $S_2$. The collision probability is simply
\begin{align}\notag
\mathbf{Pr}\left(\text{min}({\pi}(S_1)) = \text{min}({\pi}(S_2)) \right) = \frac{|S_1
  \cap S_2|}{|S_1 \cup S_2|}=R.
\end{align}
One can repeat the permutation $k$ times: $\pi_1$, $\pi_2$, ..., $\pi_k$ to estimate $R$ without bias, as
 \begin{align}\label{eqn_RM}
&\hat{R}_{M} = \frac{1}{k}\sum_{j=1}^{k}1\{{\min}({\pi_j}(S_1)) =
  {\min}({\pi_j}(S_2))\}, \\\label{eqn_Var_M}
&\text{Var}\left(\hat{R}_{M}\right) = \frac{1}{k}R(1-R).
\end{align}

The common practice of minwise hashing is to store each hashed value, e.g., ${\min}({\pi}(S_1))$ and ${\min}({\pi}(S_2))$, using 64 bits~\cite{Proc:Fetterly_WWW03}. The storage (and computational) cost  will be prohibitive in  truly large-scale (industry) applications~\cite{Proc:Manku_WWW07}.

In order to apply minwise hashing for efficiently training linear learning algorithms such as logistic regression or linear SVM, we need to express the estimator (\ref{eqn_RM}) as an inner product. For simplicity, we introduce
\begin{align}\notag
z_1 = {\min}({\pi_j}(S_1)), \hspace{0.5in} z_2 = {\min}({\pi_j}(S_2)),
\end{align}
and we hope that the term $1\{z_1 = z_2\}$  can be expressed  as an inner product. Indeed, because
\begin{align}\notag
1\{z_1 = z_2\} = \sum_{t=0}^{D-1}1\{z_1 = t\}\times 1\{z_2 = t\}
\end{align}
we know immediately that the estimator (\ref{eqn_RM}) for minwise hashing is an inner product between two extremely high-dimensional ($D\times k$) vectors. Each vector, which has  exactly $k$ 1's, is a concatenation of $k$ $D$-dimensional vectors. Because $D=2^{64}$ is possible in industry applications, the total indexing space ($D\times k$) may be too high to directly use this representation for training.\\

The recent development of {\em b-bit minwise hashing}~\cite{Proc:Li_Konig_WWW10,Proc:Li_Konig_NIPS10,Article:Li_Konig_CACM11} provides a strikingly simple solution  by storing only the lowest $b$ bits (instead of 64 bits) of each hashed value. For convenience, we define
\begin{align}\notag
e_{1,i} = i\text{th lowest bit of }z_1, \hspace{0.5in} e_{2,i} = i\text{th lowest bit of }z_2.
\end{align}

\begin{theorem}\cite{Proc:Li_Konig_WWW10}\label{The_basic}
Assume $D$ is large (i.e., $D\rightarrow\infty$).
\begin{align}\label{eqn_basic}
&P_b=\mathbf{Pr}\left(\prod_{i=1}^b1\left\{e_{1,i} = e_{2,i}\right\}\right) = C_{1,b} + \left(1-C_{2,b}\right) R,\\\notag
&r_1 = \frac{f_1}{D}, \hspace{0.1in} r_2 = \frac{f_2}{D}, \ \ f_1 = |S_1|,\  \ f_2 =|S_2|,\\\notag
&C_{1,b} = A_{1,b} \frac{r_2}{r_1+r_2} + A_{2,b}\frac{r_1}{r_1+r_2},\hspace{0.3in} \ C_{2,b} = A_{1,b} \frac{r_1}{r_1+r_2} + A_{2,b}\frac{r_2}{r_1+r_2},\\\notag
&A_{1,b} = \frac{r_1\left[1-r_1\right]^{2^b-1}}{1-\left[1-r_1\right]^{2^b}},\hspace{1in}
A_{2,b} = \frac{r_2\left[1-r_2\right]^{2^b-1}}{1-\left[1-r_2\right]^{2^b}}.\Box
\end{align}
As $r_1\rightarrow0$ and $r_2\rightarrow 0$, the limits are
\begin{align}
&A_{1,b} = A_{2,b} = C_{1,b} = C_{2,b} = \frac{1}{2^b}\\\label{eqn_Pb_r->0}
&P_b = \frac{1}{2^b} + \left(1-\frac{1}{2^b}\right)R
\end{align}
\end{theorem}

The case $r_1, r_2\rightarrow 0$ is very common in practice because the data are often relatively highly sparse (i.e., $r_1, r_2\approx 0$), although they can be very large in the absolute scale. For example, if $D=2^{64}$, then a set $S_1$ with $f_1=|S_1| = 2^{54}$ (which roughly corresponds to the size of a small novel) is highly sparse ($r_1 \approx 0.001$) even though $2^{54}$ is actually very large in the absolute scale. One can also verify that the error by using (\ref{eqn_Pb_r->0}) to replace (\ref{eqn_basic}) is bounded by $O(r_1 + r_2)$, which is very small when $r_1, r_2\rightarrow 0$. In fact, \cite{Proc:Li_Konig_NIPS10} extensively used this argument for studying 3-way set similarities.

We can then estimate $P_b$ (and $R$) from $k$ independent permutations: $\pi_1$, $\pi_2$, ..., $\pi_k$,
\begin{align}\label{eqn_R_b}
&\hat{R}_b = \frac{\hat{P}_b - C_{1,b}}{1-C_{2,b}},\hspace{0.6in} \hat{P}_{b} = \frac{1}{k}\sum_{j=1}^{k}\left\{ \prod_{i=1}^b1\{e_{1,i,\pi_j} = e_{2,i,\pi_j}\}\right\},\\\label{eqn_Var_b}
&\text{Var}\left(\hat{R}_b\right) = \frac{\text{Var}\left(\hat{P}_b\right)}{\left[1-C_{2,b}\right]^2}
=\frac{1}{k}\frac{\left[C_{1,b}+(1-C_{2,b})R\right]\left[1-C_{1,b}-(1-C_{2,b})R\right]}{\left[1-C_{2,b}\right]^2}
\end{align}

Clearly, the similarity ($R$) information  is adequately encoded in $P_b$. In other words, often there is no need to explicitly estimate $R$.  The estimator $\hat{P}_b$ is an inner product between two vectors in $2^b\times k$ dimensions with exactly $k$ 1's. Therefore,
if $b$ is not too large (e.g., $b\leq 16$), this intuition provides a simple practical strategy for using $b$-bit minwise hashing for large-scale learning.

\section{Integrating  $b$-Bit Minwise Hashing with (Linear) Learning Algorithms}

Linear algorithms such as linear SVM and logistic regression have become very powerful and extremely popular. Representative software packages include SVM$^\text{perf}$~\cite{Proc:Joachims_KDD06}, Pegasos~\cite{Proc:Shalev-Shwartz_ICML07}, Bottou's SGD SVM~\cite{URL:Bottou_SGD}, and LIBLINEAR~\cite{Article:Fan_JMLR08}.

Given a dataset $\{(\mathbf{x}_i, y_i)\}_{i=1}^n$, $\mathbf{x}_i\in\mathbb{R}^{D}$, $y_i\in\{-1,1\}$, the $L_2$-regularized linear SVM solves the following optimization  problem:
\begin{align}
\min_{\mathbf{w}}\ \ \frac{1}{2}\mathbf{w^Tw} + C \sum_{i=1}^n \max \left\{1 - y_i\mathbf{w^Tx_i},\ 0\right\},
\end{align}
and the $L_2$-regularized logistic regression solves a  similar problem:
\begin{align}
\min_{\mathbf{w}}\ \ \frac{1}{2}\mathbf{w^Tw} + C \sum_{i=1}^n \log \left(1 + e^{-y_i\mathbf{w^Tx_i}}\right).
\end{align}
Here $C>0$ is an important penalty parameter. Since our purpose is to demonstrate the effectiveness of our proposed scheme using $b$-bit hashing, we simply provide results for a wide range of $C$ values and  assume that the best performance is achievable if we conduct cross-validations.

In our approach, we apply $k$ independent random permutations on each feature vector $\mathbf{x}_i$ and store the lowest $b$ bits of each hashed value. This way, we obtain a new dataset which can be stored using merely $nbk$ bits. At run-time,  we expand each new data point into a $2^b\times k$-length vector.\\

For example, suppose $k=3$ and the  hashed values are  originally $\{12013, 25964, 20191\}$, whose binary digits are $\{010111011101101, \ 110010101101100, 100111011011111\}$. Consider $b=2$. Then the binary digits are stored as $\{01, 00, 11\}$ (which corresponds to $\{1, 0, 3\}$ in decimals). At run-time, we need to expand them into a vector of length $2^bk = 12$, to be $\{ 0, 0, 1, 0, \ \ 0, 0, 0, 1, \ \ 1, 0, 0, 0\}$,
which will be the new feature vector fed to a solver:
\begin{align}\notag
\begin{array}{lrrr}
\text{Original hashed values } (k=3): &12013 &25964 &20191\\
\text{Original binary representations}: &010111011101101& 110010101101100& 100111011011111\\
\text{Lowest $b=2$ binary digits}: &01& 00& 11\\
\text{Expanded $2^b=4$ binary digits }: &0 0 1 0 & 0 0 0 1 & 1 0 0 0\\
\text{New feature vector fed to a solver}: &&\{0, 0, 1, 0, 0, 0, 0, 1, 1, 0, 0, 0\}
\end{array}
\end{align}

\section{Experimental Results of $b$-Bit Minwise Hashing on  Expanded RCV1 Dataset}

In our earlier technical reports~\cite{Report:Li_Moore_Konig_HashingSVM,Report:HashLearning11}, our experimental settings closely followed the work in~\cite{Proc:Yu_KDD10} by testing $b$-bit minwise hashing on the {\em webspam} dataset ($n=350000$, $D=16609143$). Following~\cite{Proc:Yu_KDD10}, we randomly selected $20\%$ of samples for testing and used the remaining $80\%$ samples for training. Since the {\em webspam} dataset (24 GB in LibSVM format) may be too small compared by datasets used in industry, in this paper we present an empirical study on the {\em expanded rcv1} dataset by using the original features + all pairwise combinations (products) of features + 1/30 of 3-way combinations (products) of features.

We chose LIBLINEAR as the tool to demonstrate the effectiveness of our algorithm.  All experiments were conducted on  workstations with Xeon(R) CPU (W5590@3.33GHz) and 48GB RAM, under Windows 7 System.

\vspace{-0.1in}

\begin{table}[h]
\caption{Data information}
\begin{center}{
\begin{tabular}{l l l l l}
\hline \hline
Dataset & \# Examples ($n$) &\# Dimensions ($D$) & \# Nonzeros  Median (Mean) &Train / Test Split\\\hline
Webspam (24 GB) &350000  &16609143 & 3889 (3728) &$80\%$ / $20\%$ \\
Rcv1 (200 GB) &677399  &1010017424  &3051 (12062) &$50\%$ / $50\%$\\
\hline\hline
\end{tabular}
}
\end{center}
\label{tab_preprocessing}
\end{table}\vspace{-0.1in}

Note that we  used the original ``testing'' data of {\em rcv1} to generate our new expanded dataset. The original ``training'' data of {\em rcv1} had only 20242 examples. Also, to ensure reliable  test results, we randomly split our  expanded {\em rcv1} dataset into
two halves, for training and testing.


\subsection{ Experimental Results Using Linear SVM}

Since there is an important tuning parameter $C$ in linear SVM and logistic regression, we conducted our extensive experiments for a wide range of $C$ values (from $10^{-3}$ to $10^2$) with finer spacings in $[0.1,\ 10]$.

\begin{figure}[h!]
\mbox{
\includegraphics[width=1.7in]{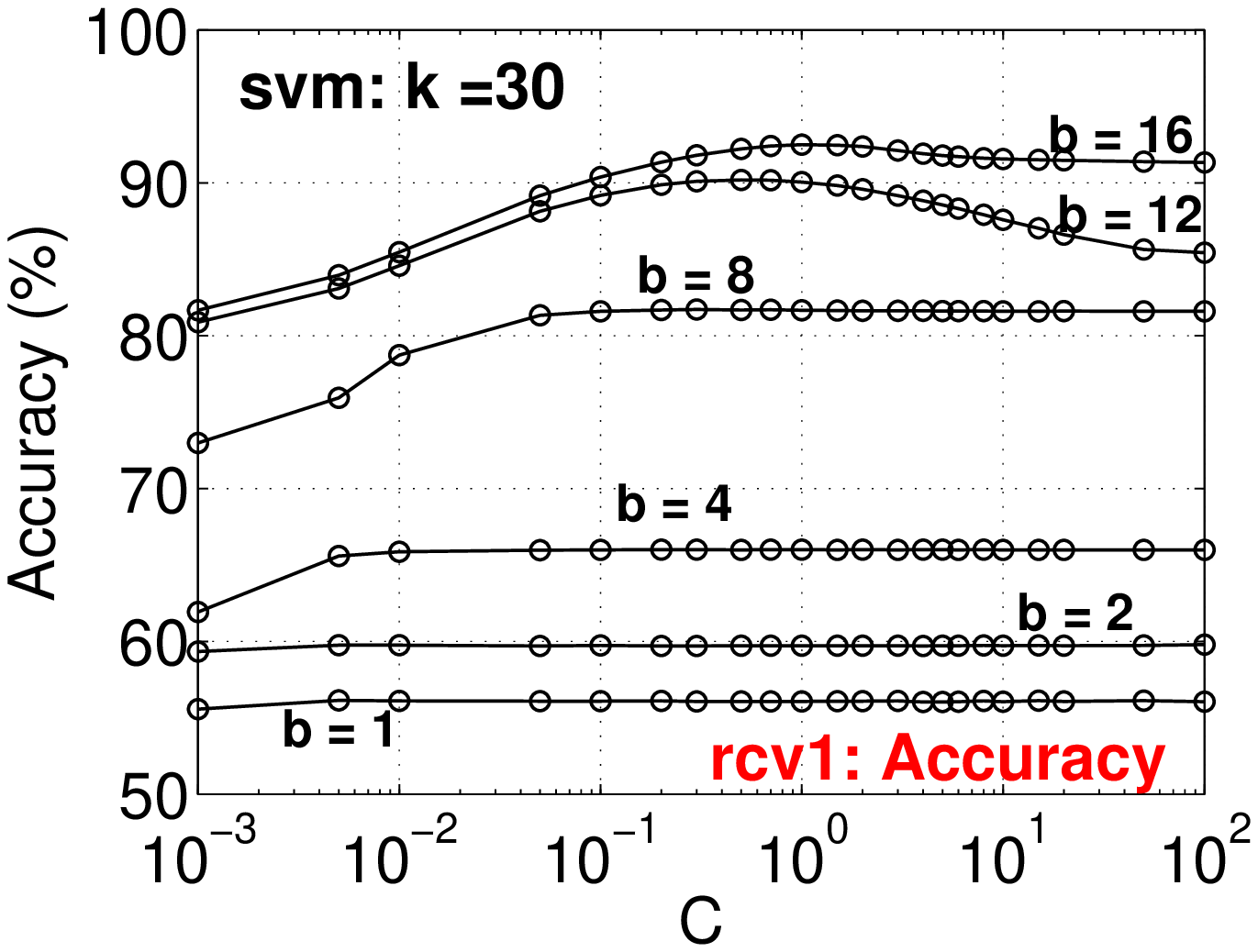}\hspace{-0.1in}
\includegraphics[width=1.7in]{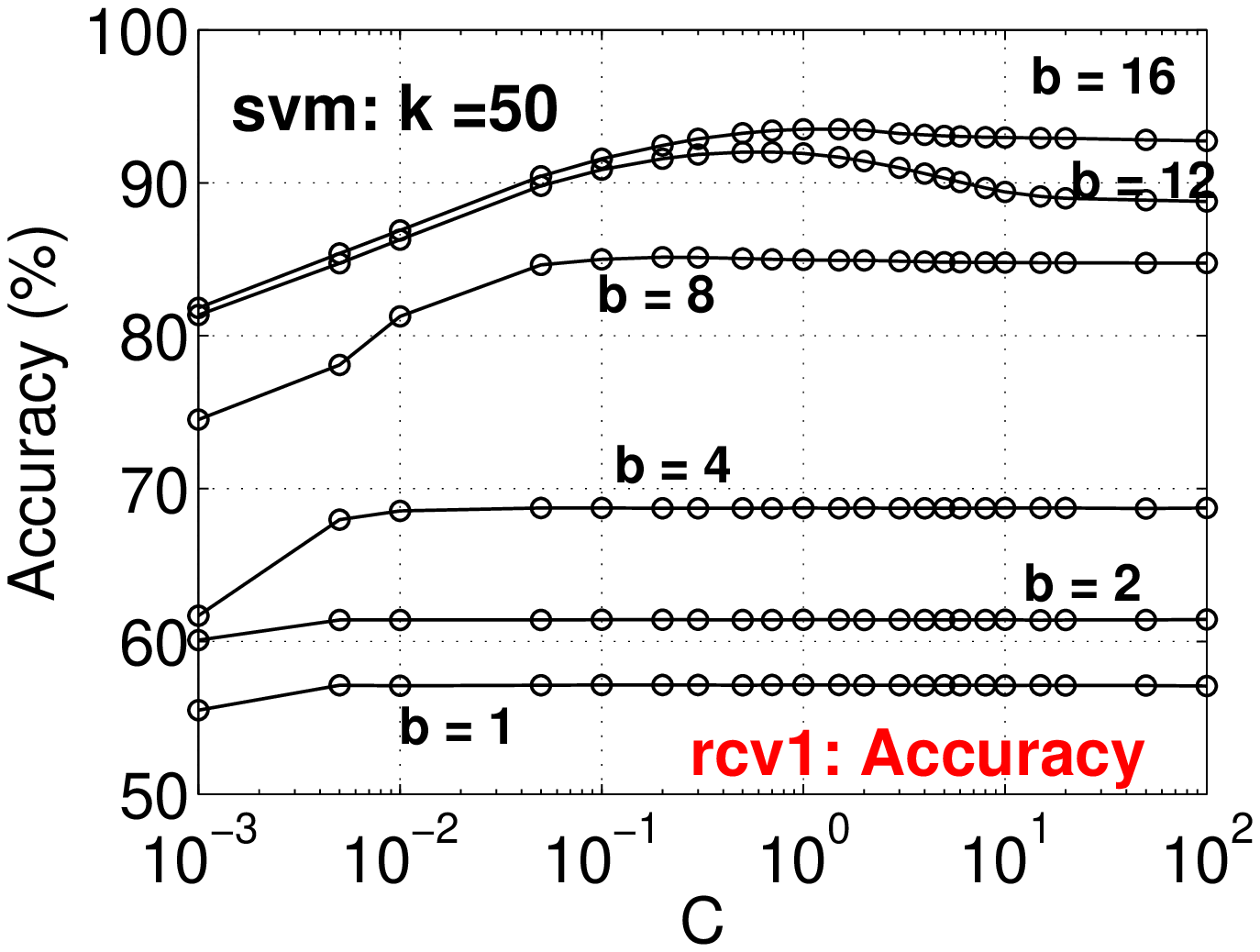}\hspace{-0.1in}
\includegraphics[width=1.7in]{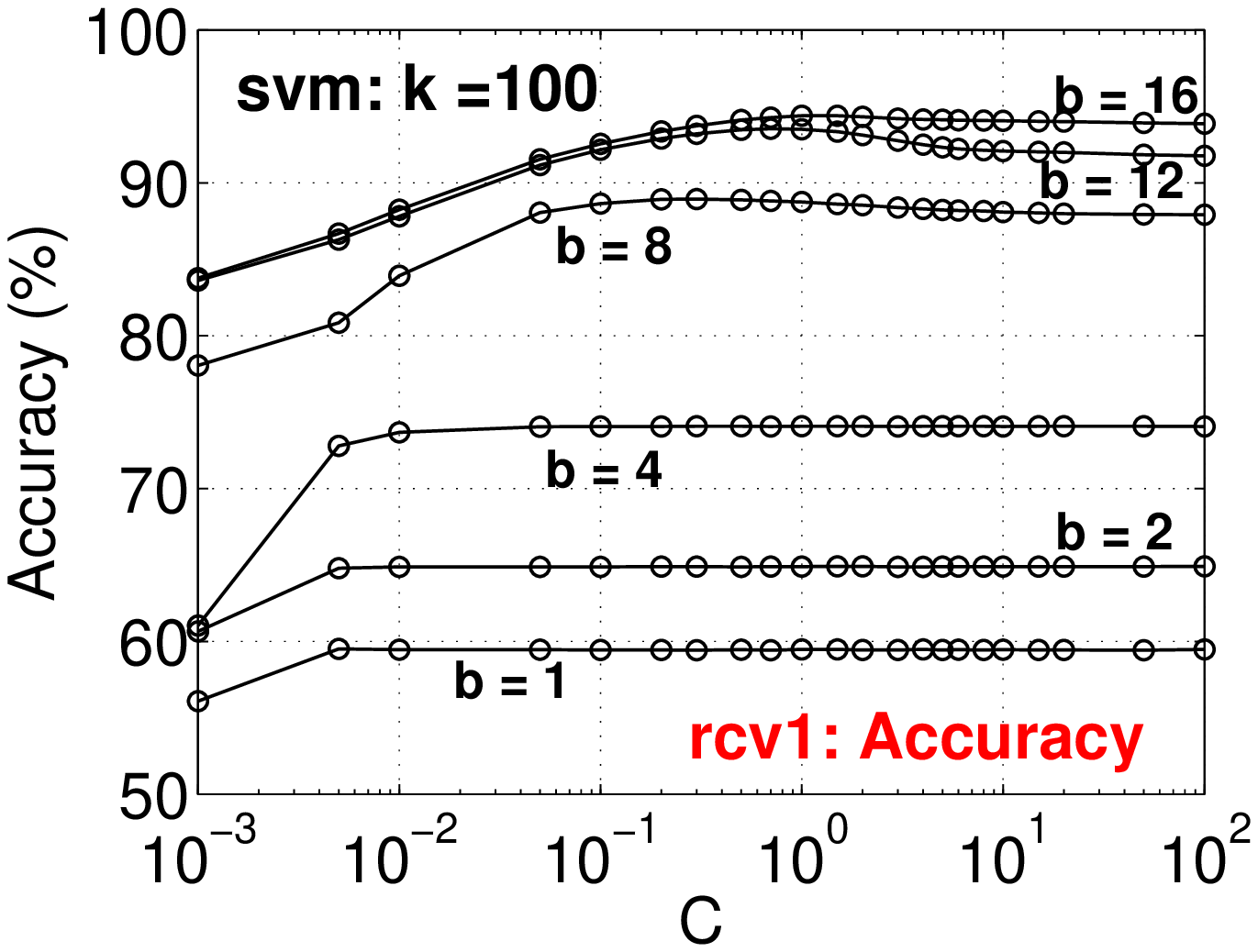}\hspace{-0.1in}
\includegraphics[width=1.7in]{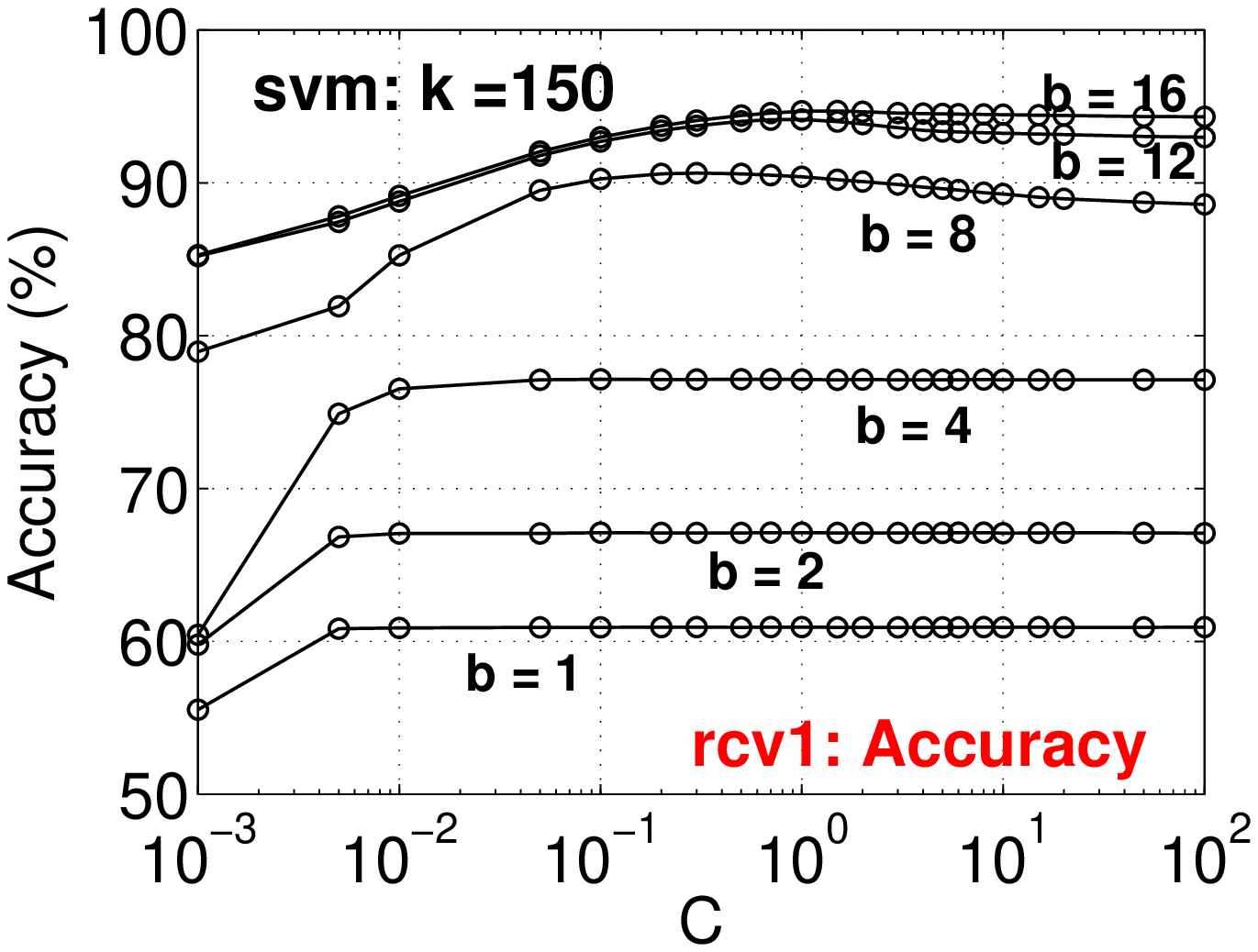}}

\mbox{
\includegraphics[width=1.7in]{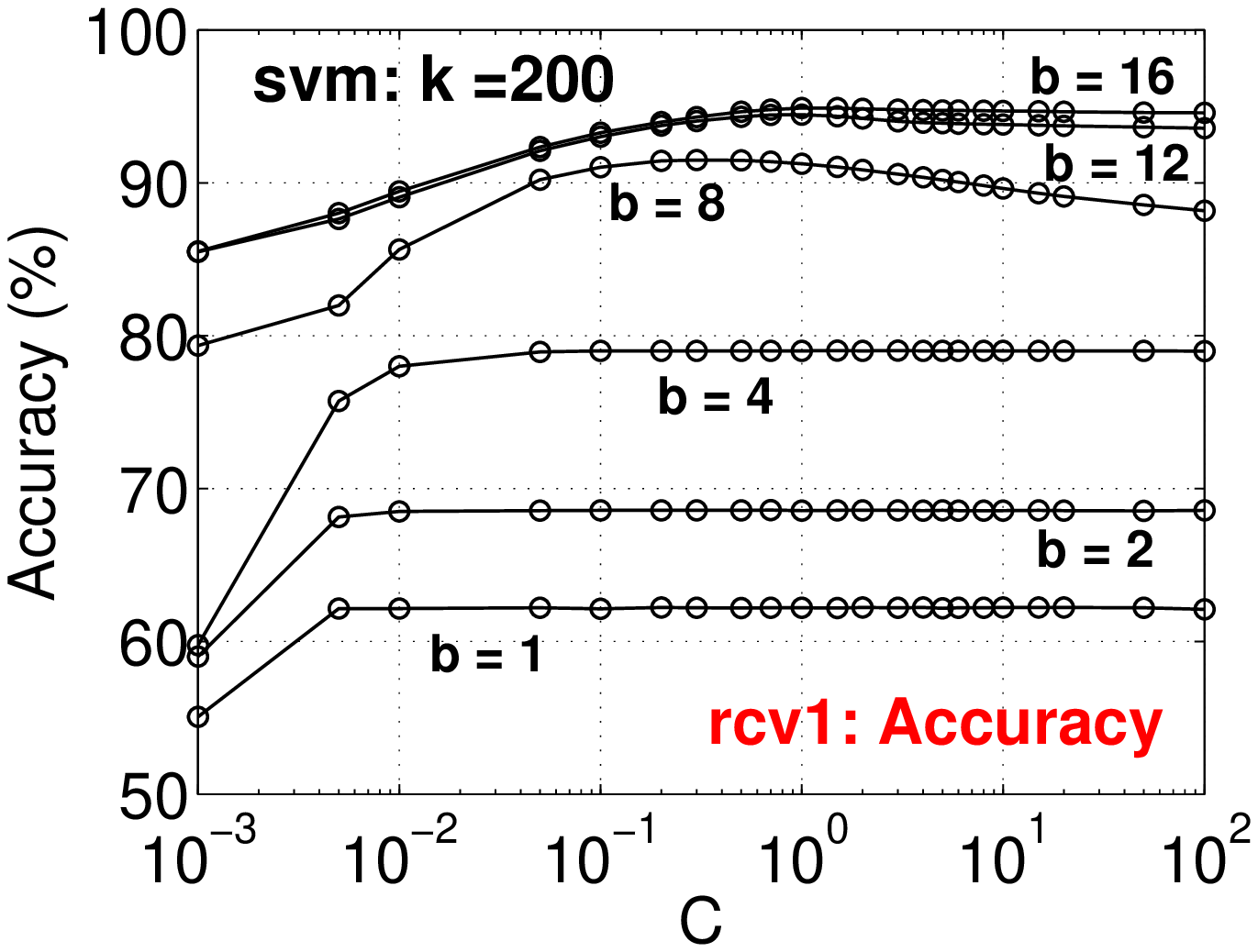}\hspace{-0.1in}
\includegraphics[width=1.7in]{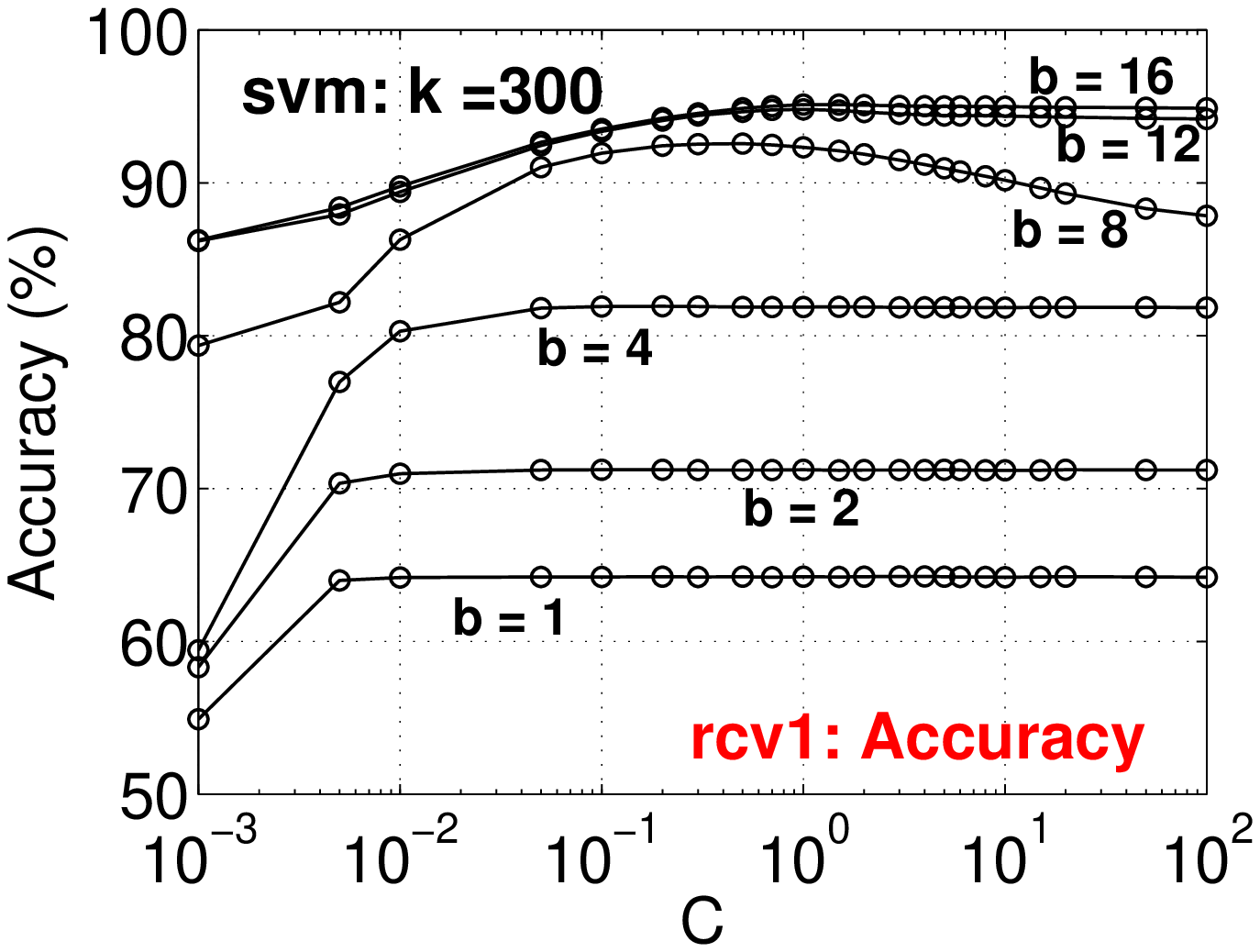}\hspace{-0.1in}
\includegraphics[width=1.7in]{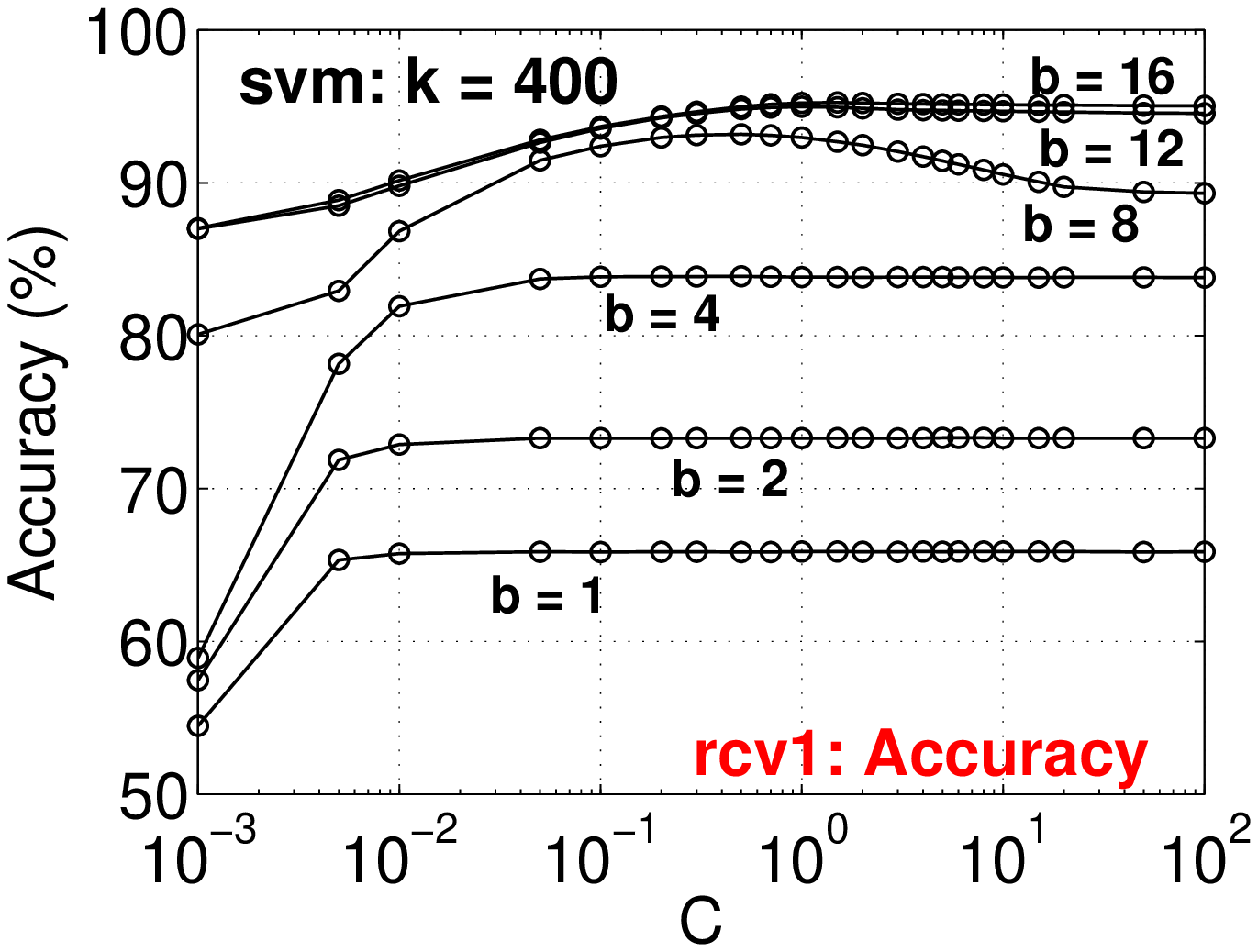}\hspace{-0.1in}
\includegraphics[width=1.7in]{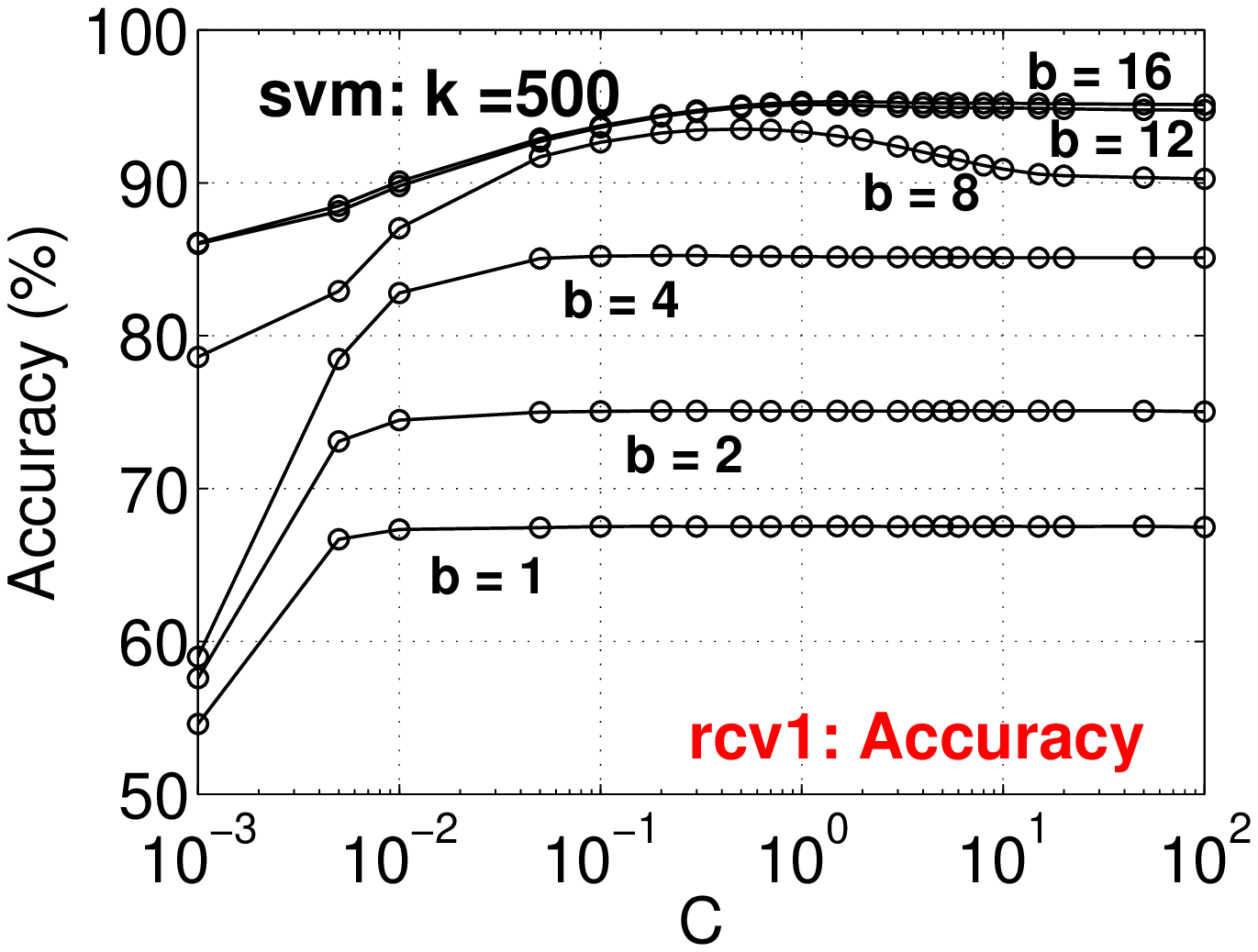}}

\vspace{-0.15in}

\caption{\textbf{Linear SVM test accuracy on rcv1}.   }\label{fig_rcv1_acc}
\end{figure}

\begin{figure}[h!]
\mbox{
\includegraphics[width=1.7in]{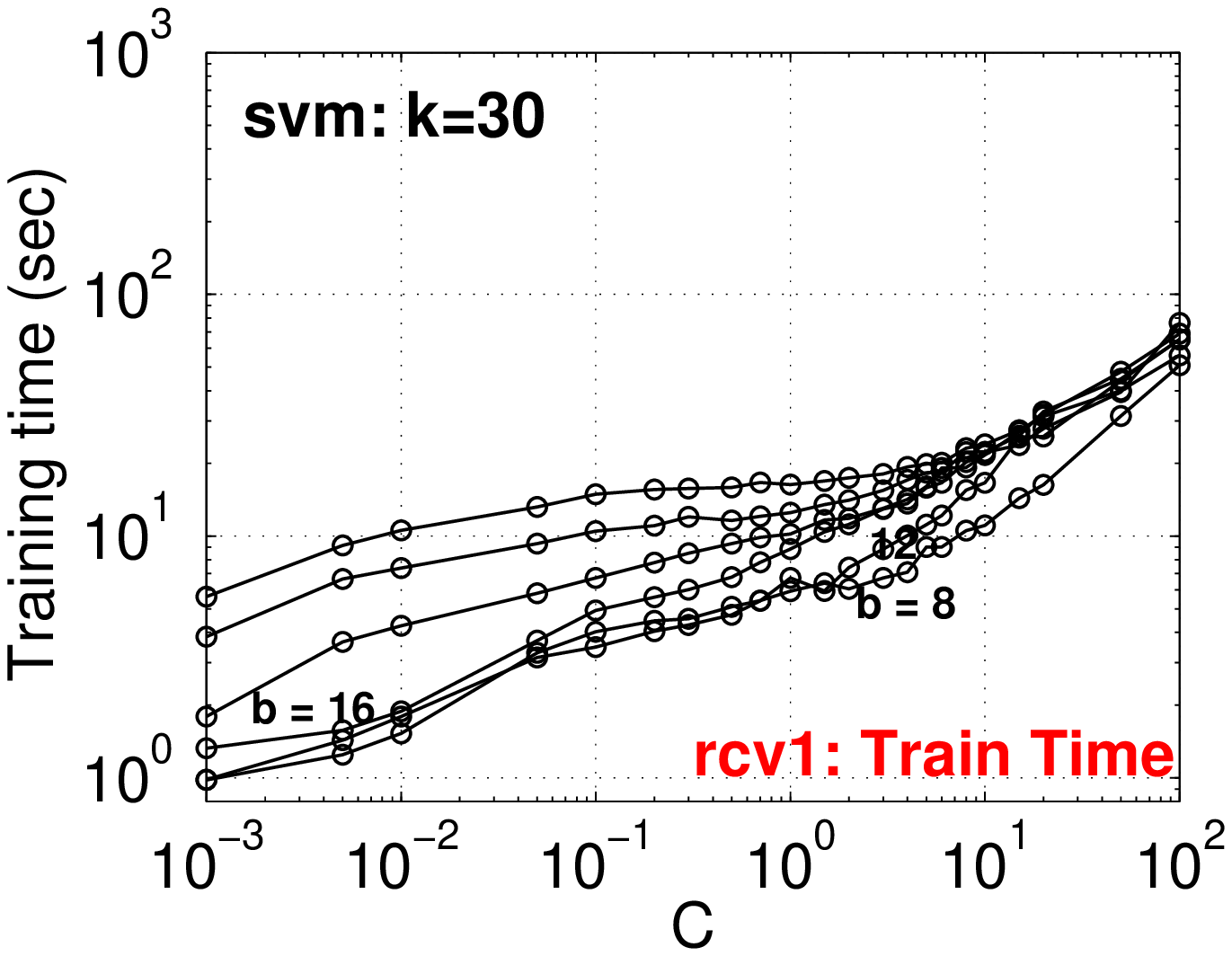}\hspace{-0.1in}
\includegraphics[width=1.7in]{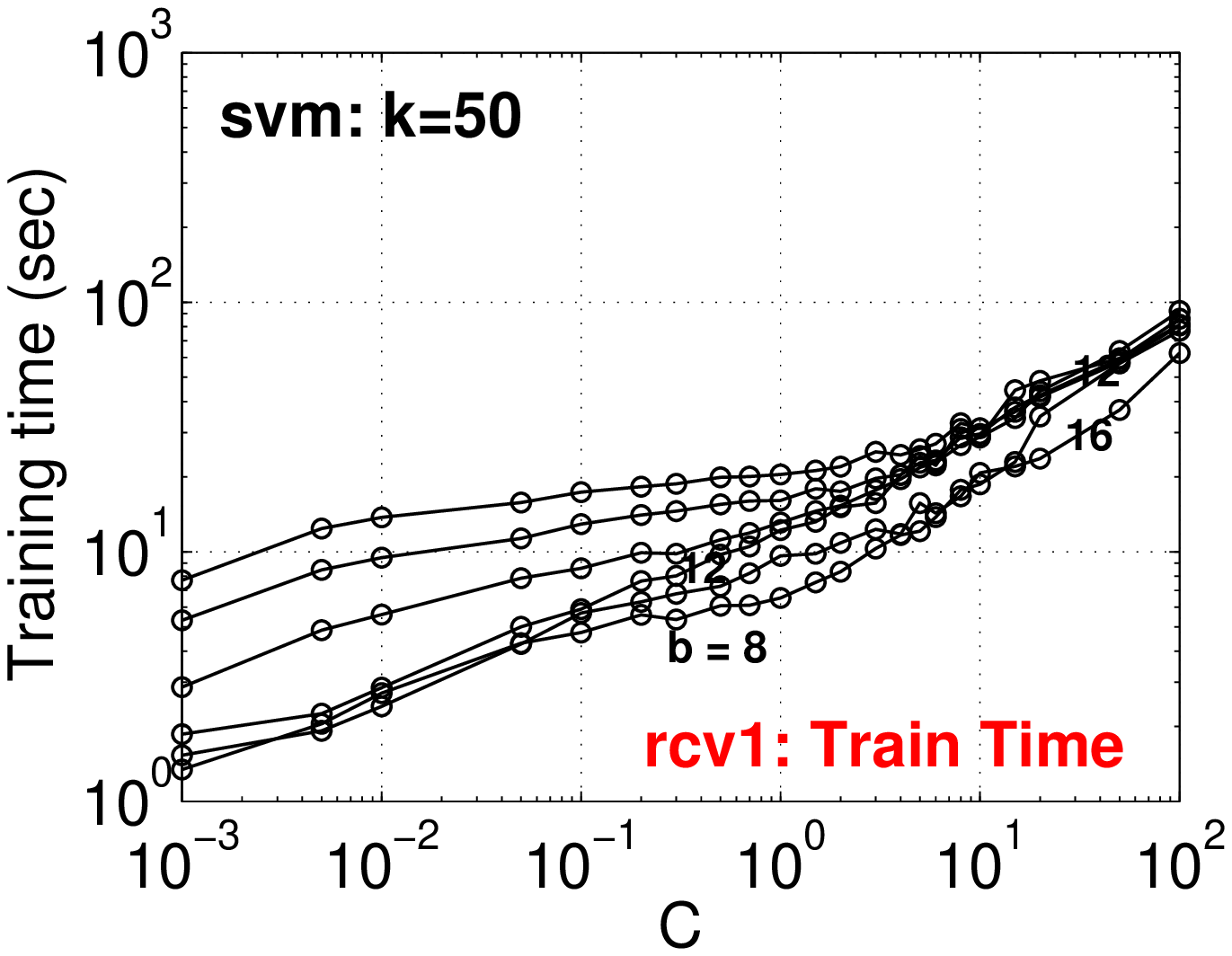}\hspace{-0.1in}
\includegraphics[width=1.7in]{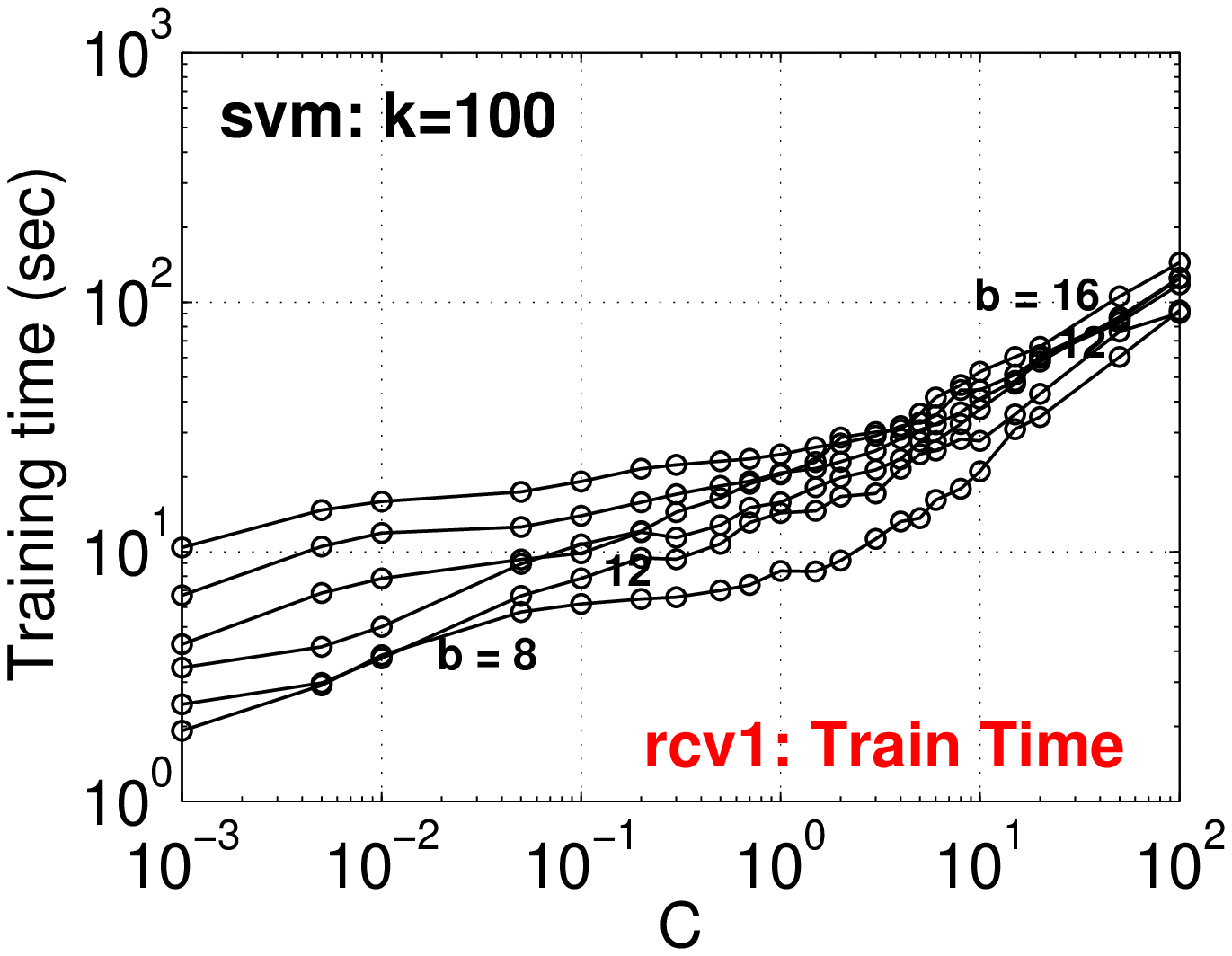}\hspace{-0.1in}
\includegraphics[width=1.7in]{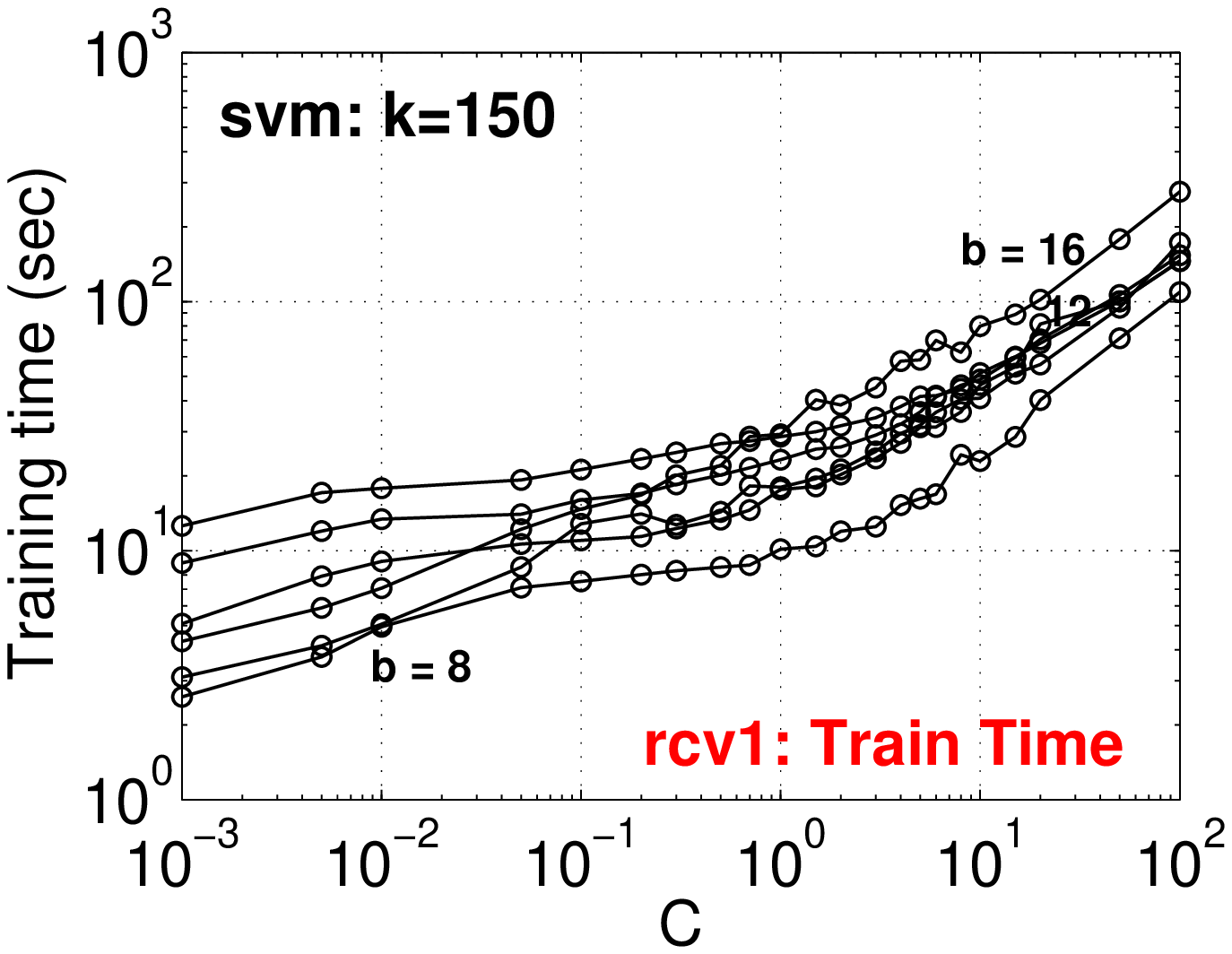}}

\mbox{
\includegraphics[width=1.7in]{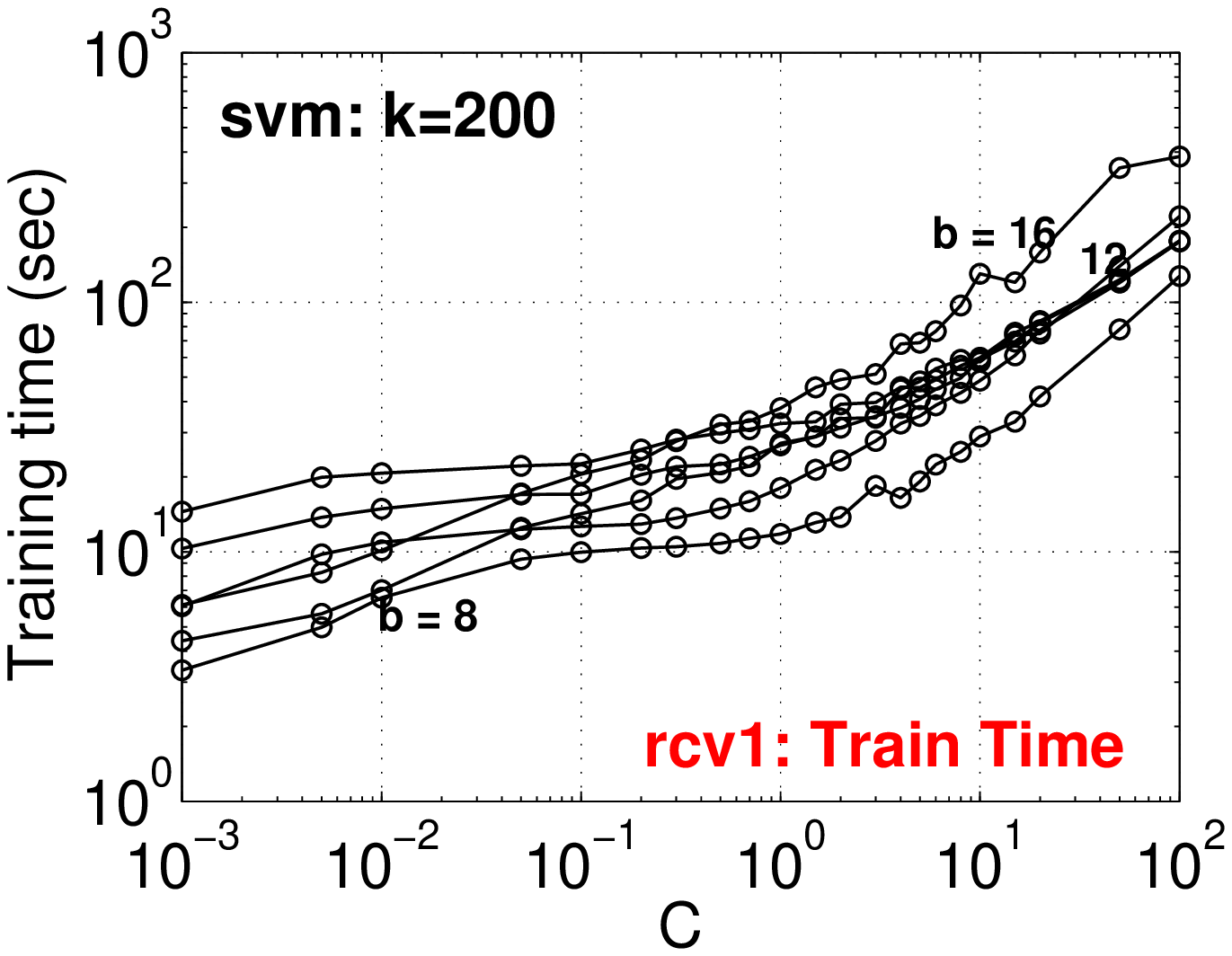}\hspace{-0.1in}
\includegraphics[width=1.7in]{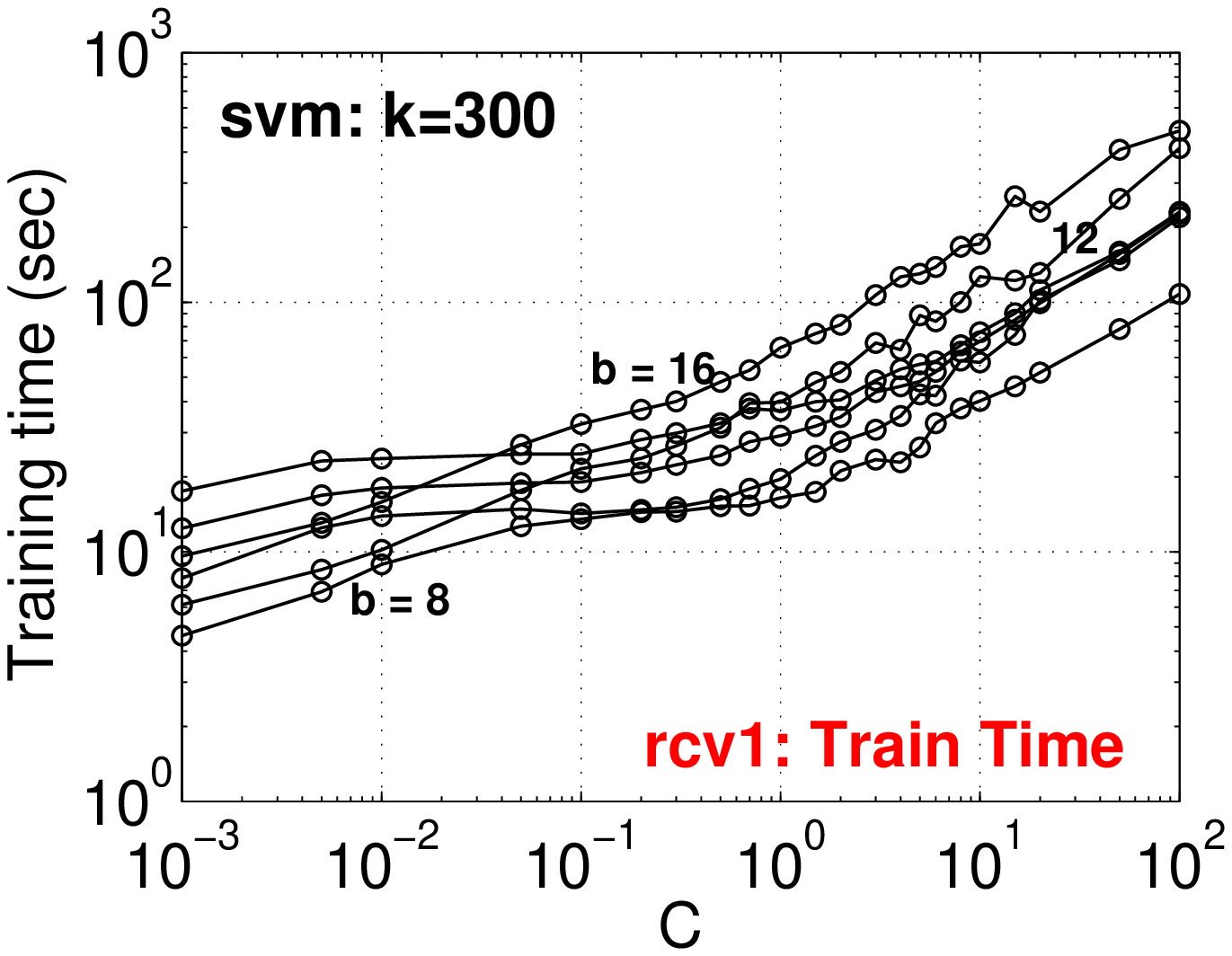}\hspace{-0.1in}
\includegraphics[width=1.7in]{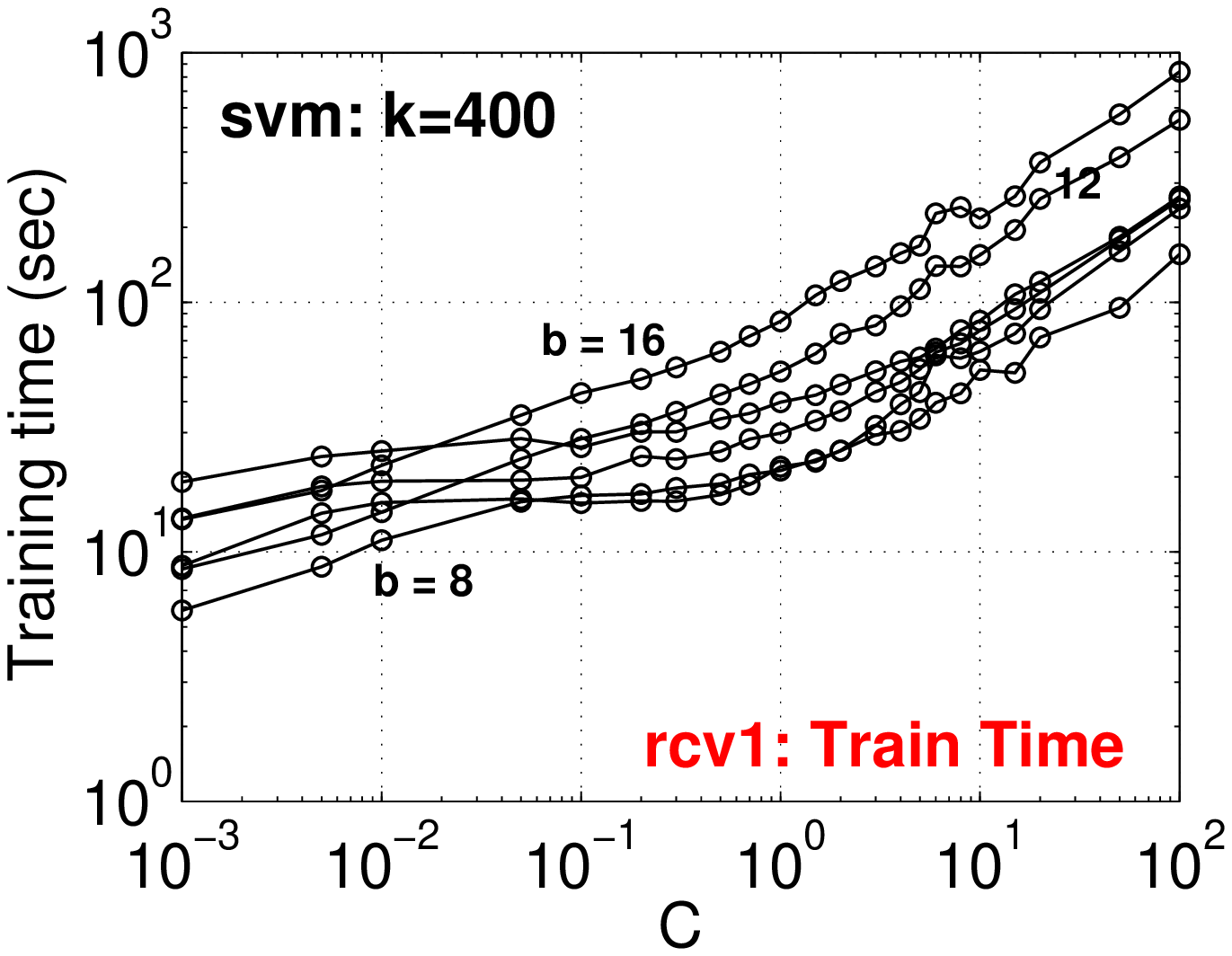}\hspace{-0.1in}
\includegraphics[width=1.7in]{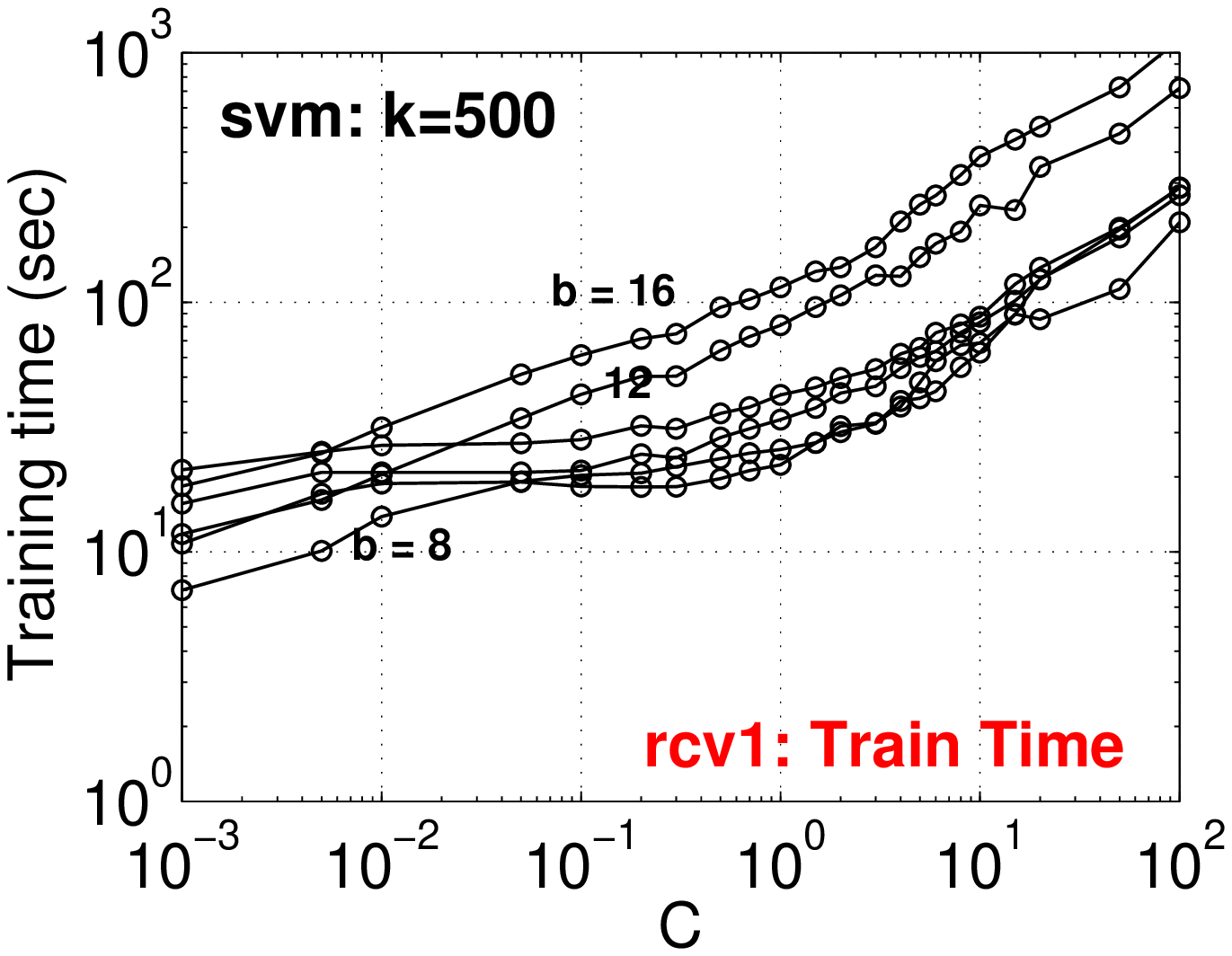}}

\vspace{-0.25in}

\caption{\textbf{Linear SVM training time on rcv1}.}\label{fig_rcv1_train}
\end{figure}

We mainly experimented with $k=30$ to $k=500$, and $b=1$, 2, 4, 8, 12, and 16. Figures~\ref{fig_rcv1_acc} and~\ref{fig_rcv1_train}   provide the test accuracies and train times, respectively.  We hope in the near future we will add the baseline results by training LIBLINEAR on the entire (200 GB) dataset, once we have the resources to do so. Note that with merely $k=30$, we can achieve $>90\%$ test accuracies (using $b=12$). The VW algorithm can also achieve $90\%$ accuracies with $k=2^{14}$.

For this dataset, the best performances were usually achieved when  $C\geq 1$. Note that we plot all the results for different $C$ values in one figure so that others can easily verify our work, to maximize the repeatability.

\vspace{-0.2in}
\subsection{ Experimental Results Using Logistic Regression}

\begin{figure}[h!]
\mbox{
\includegraphics[width=1.7in]{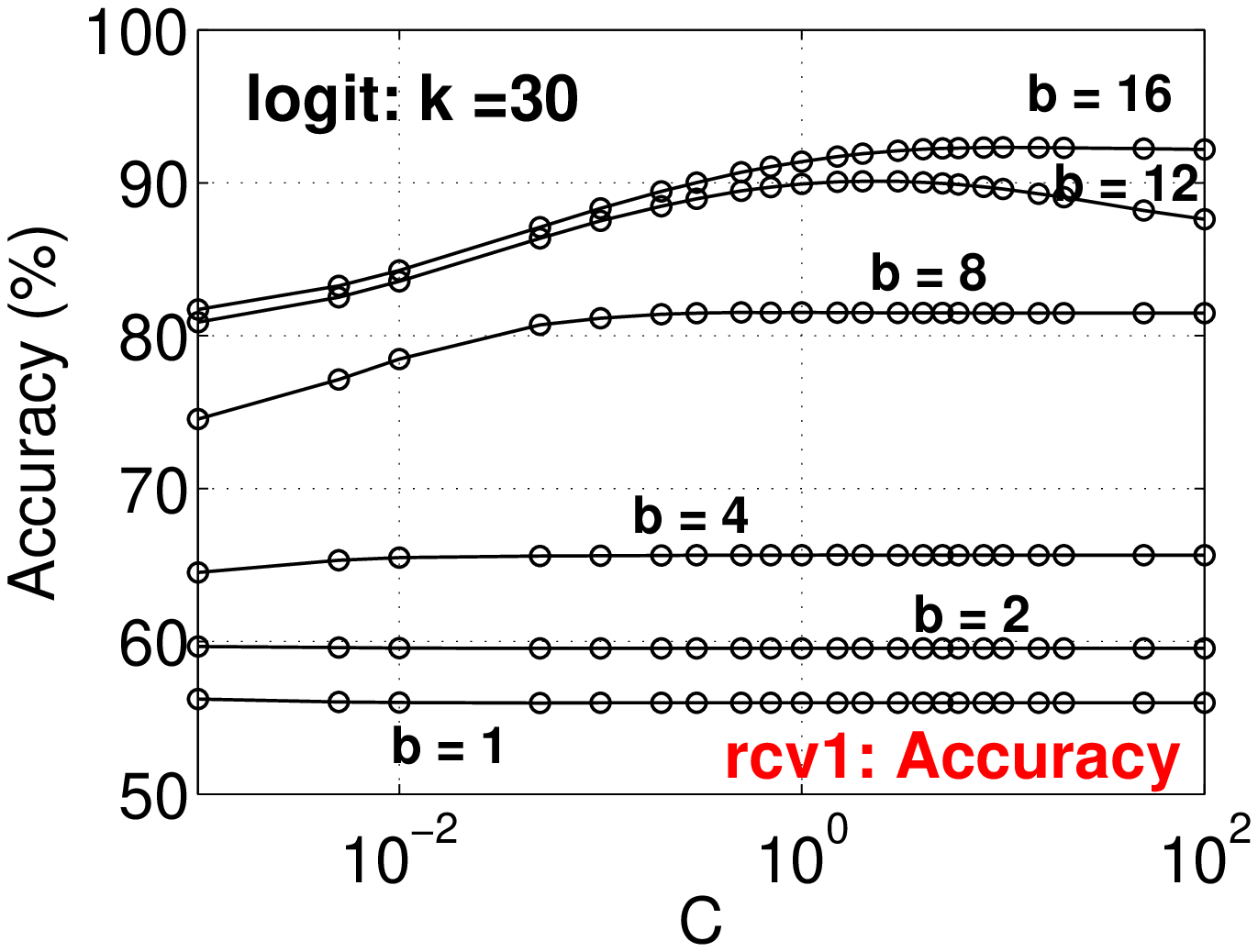}\hspace{-0.1in}
\includegraphics[width=1.7in]{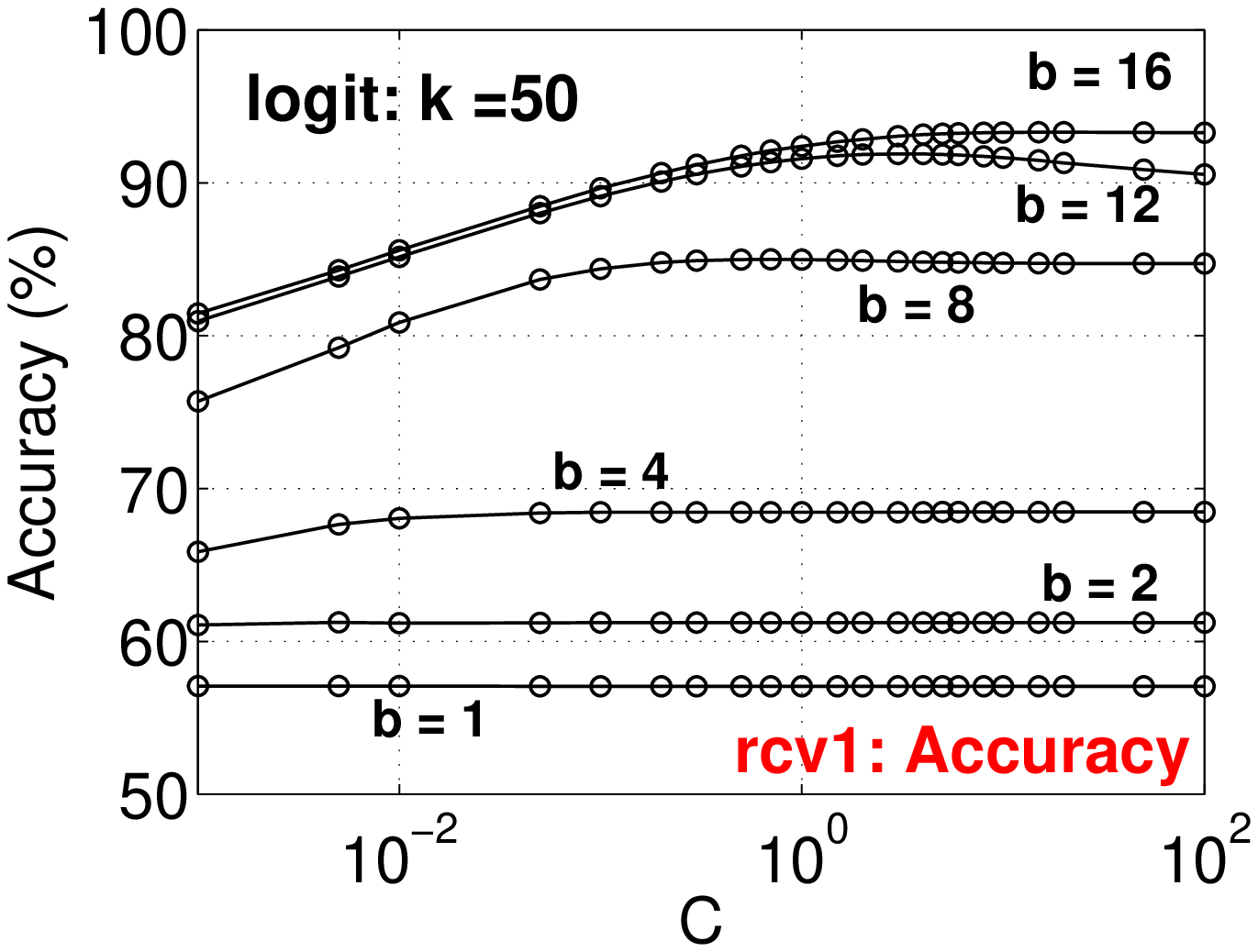}\hspace{-0.1in}
\includegraphics[width=1.7in]{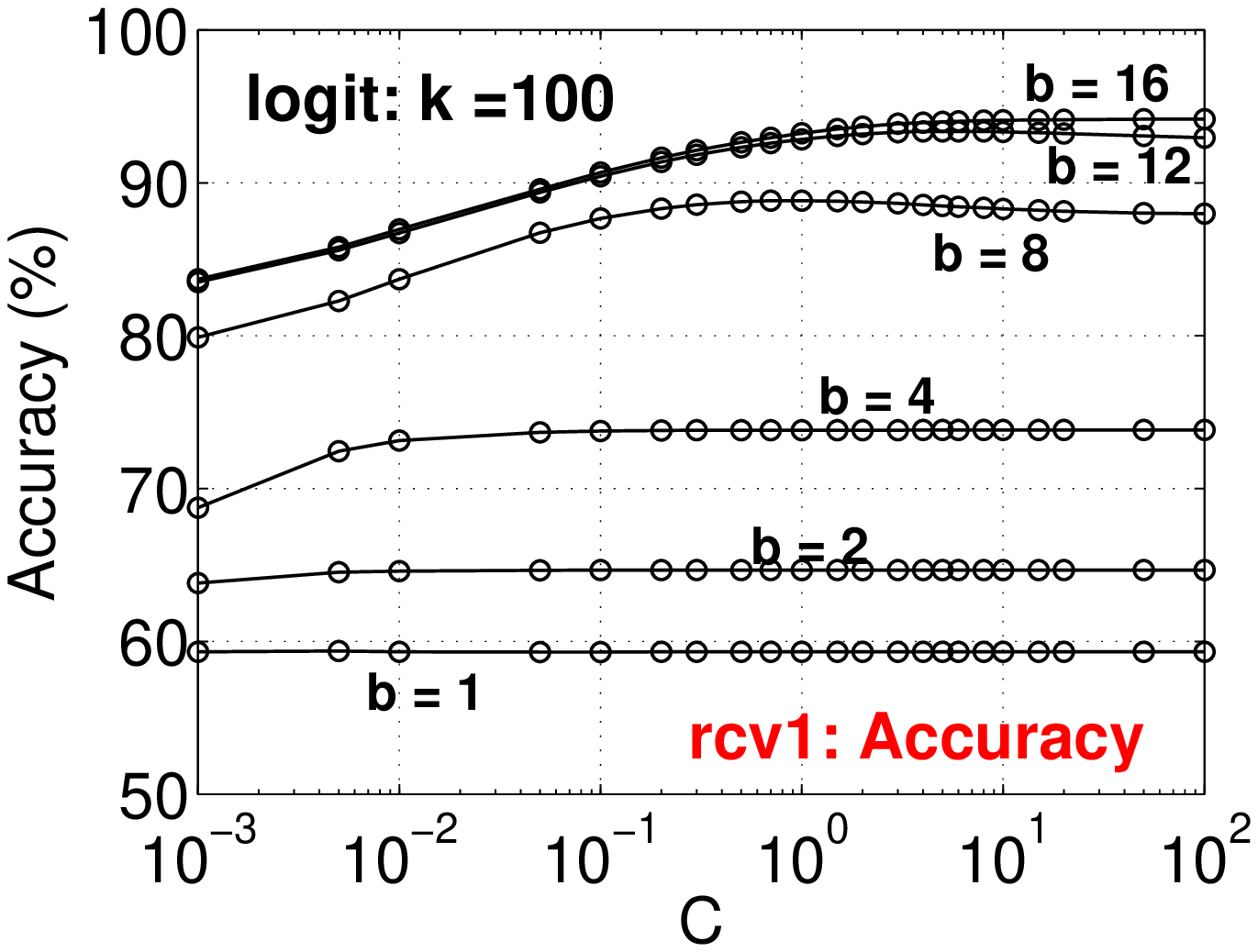}\hspace{-0.1in}
\includegraphics[width=1.7in]{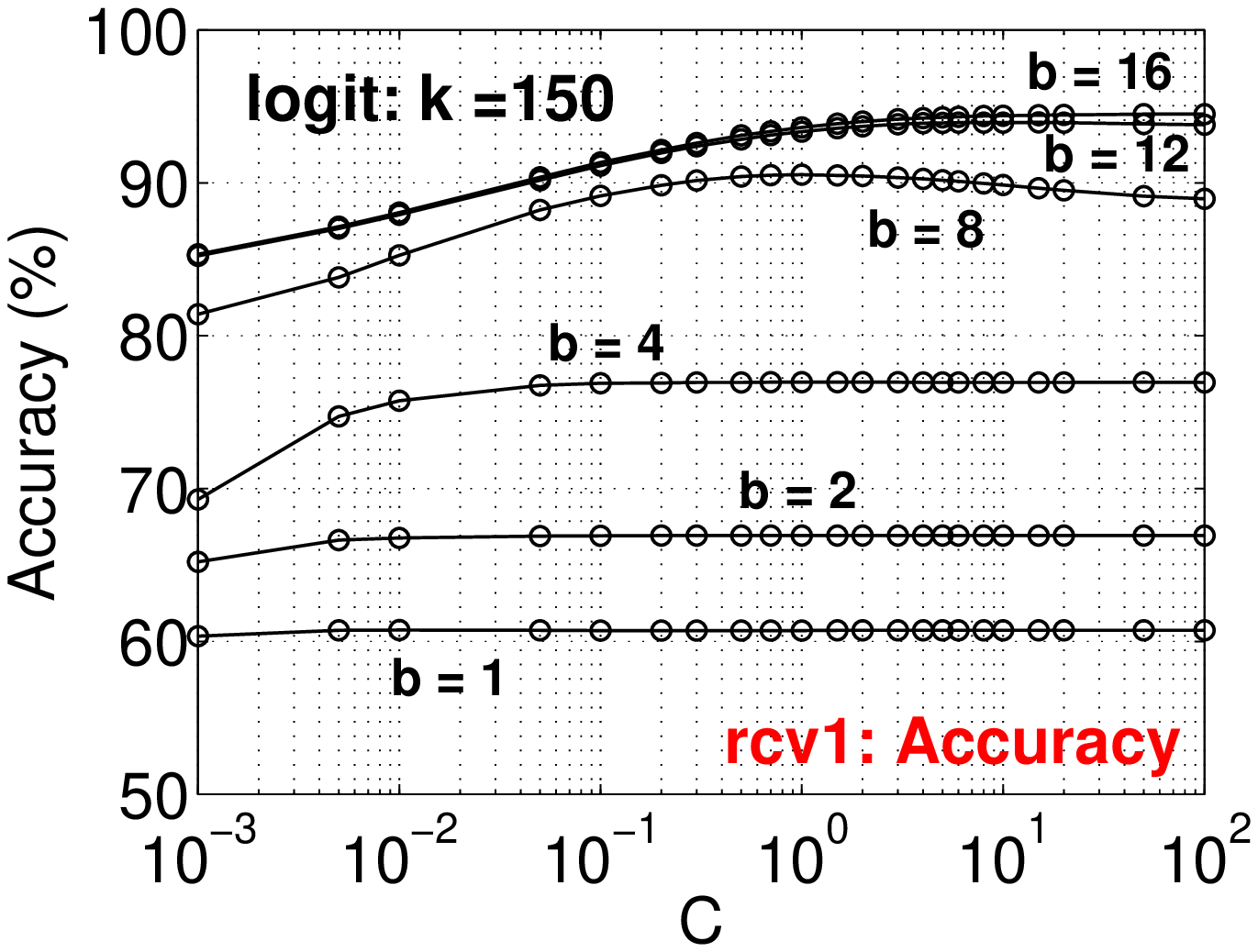}}

\mbox{
\includegraphics[width=1.7in]{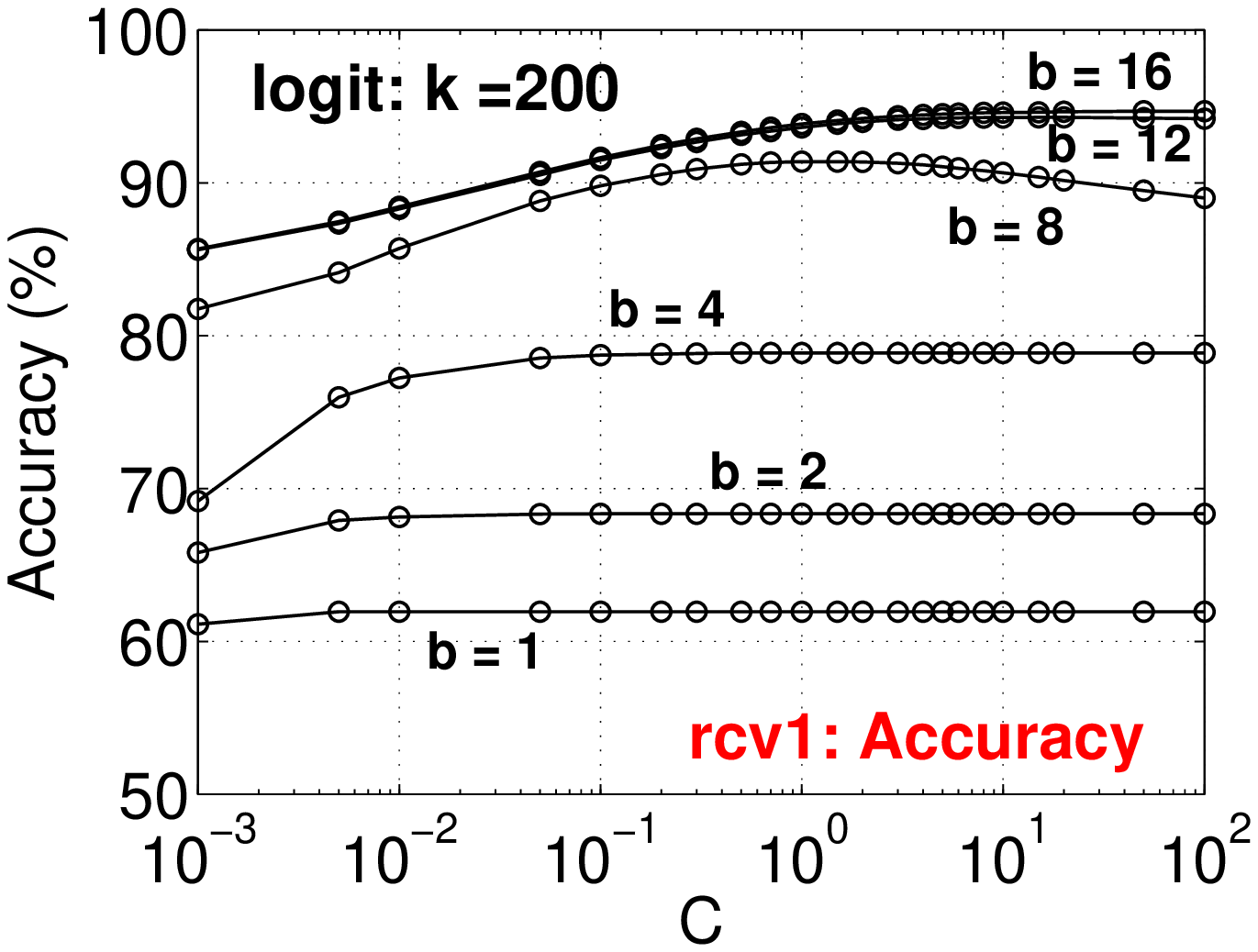}\hspace{-0.1in}
\includegraphics[width=1.7in]{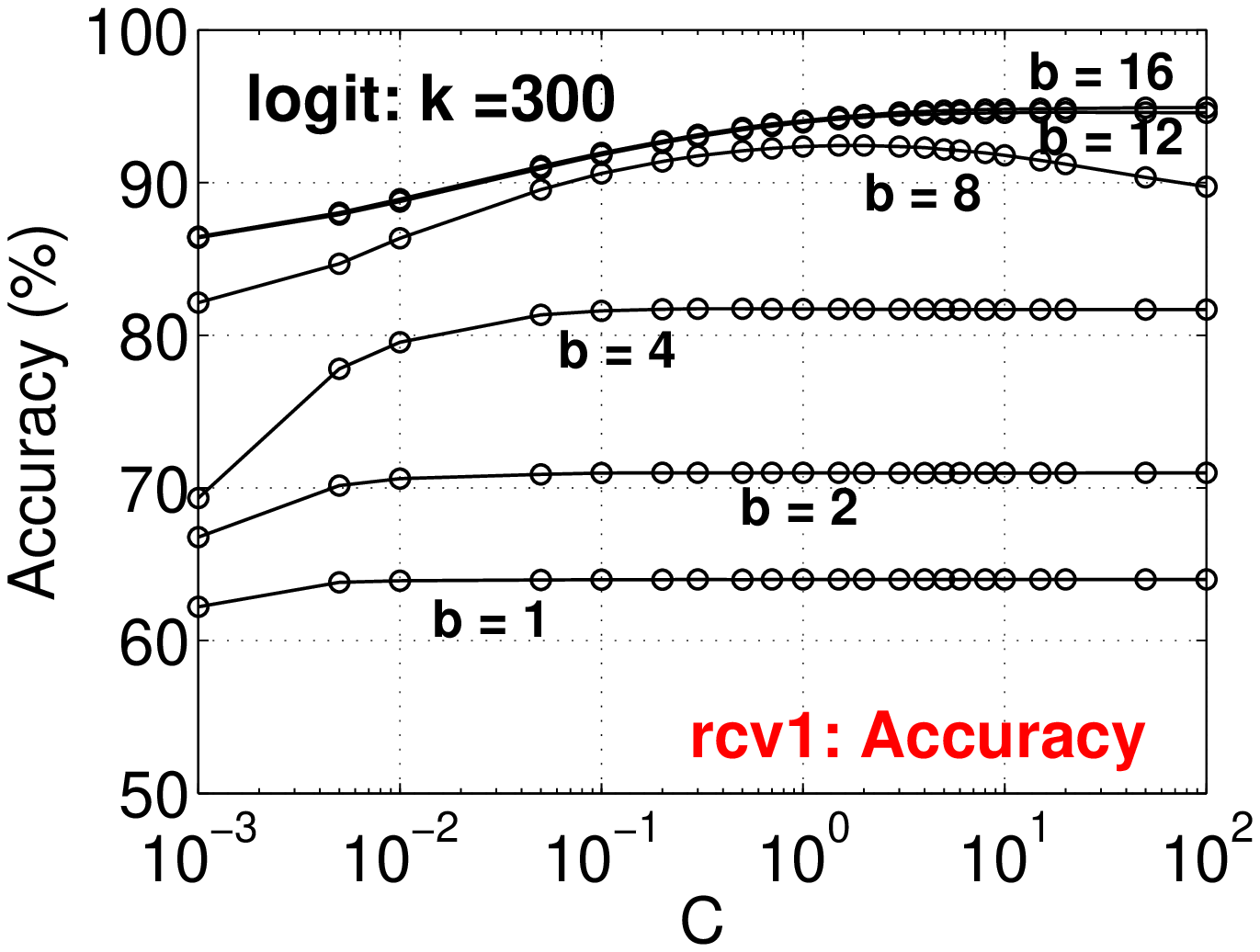}\hspace{-0.1in}
\includegraphics[width=1.7in]{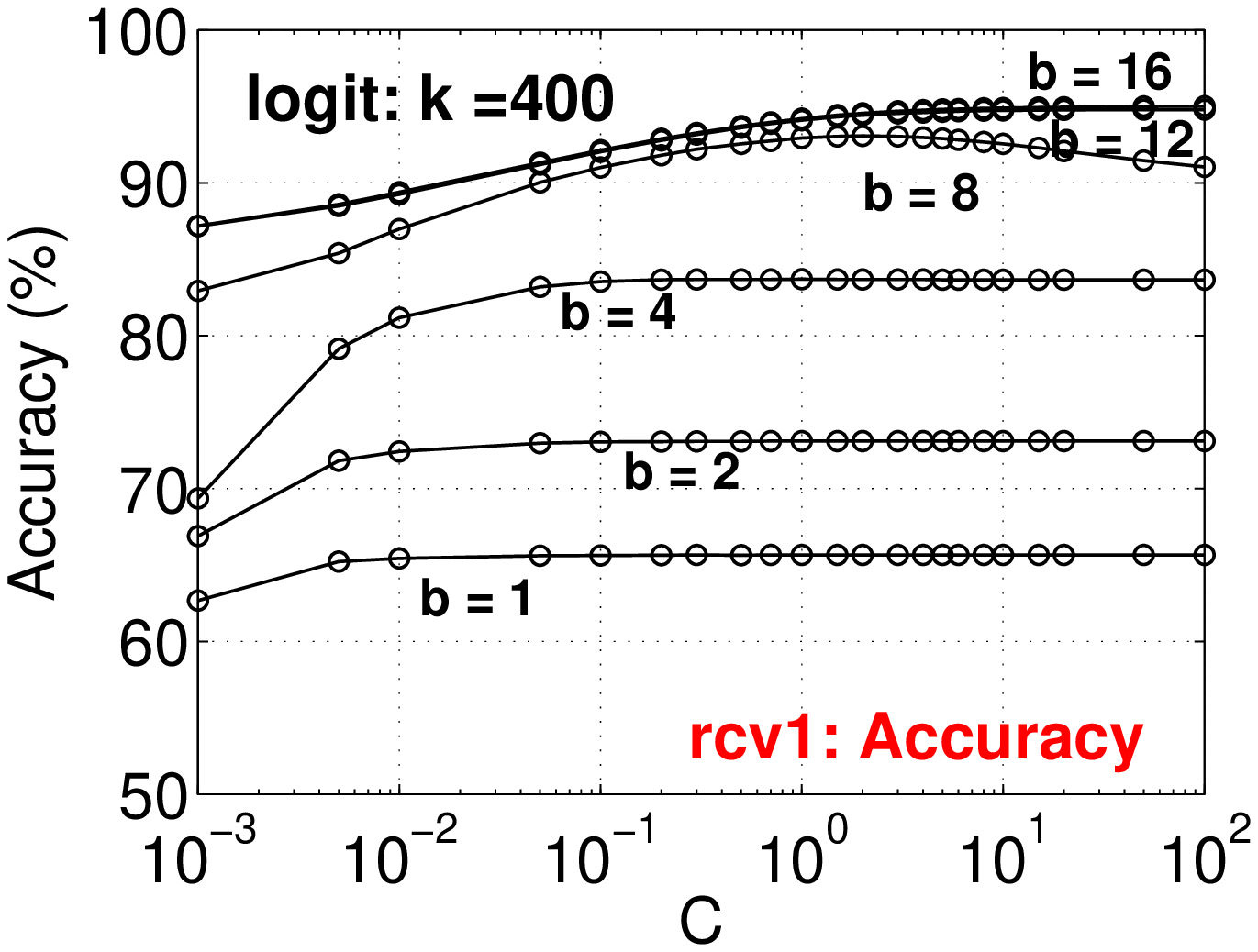}\hspace{-0.1in}
\includegraphics[width=1.7in]{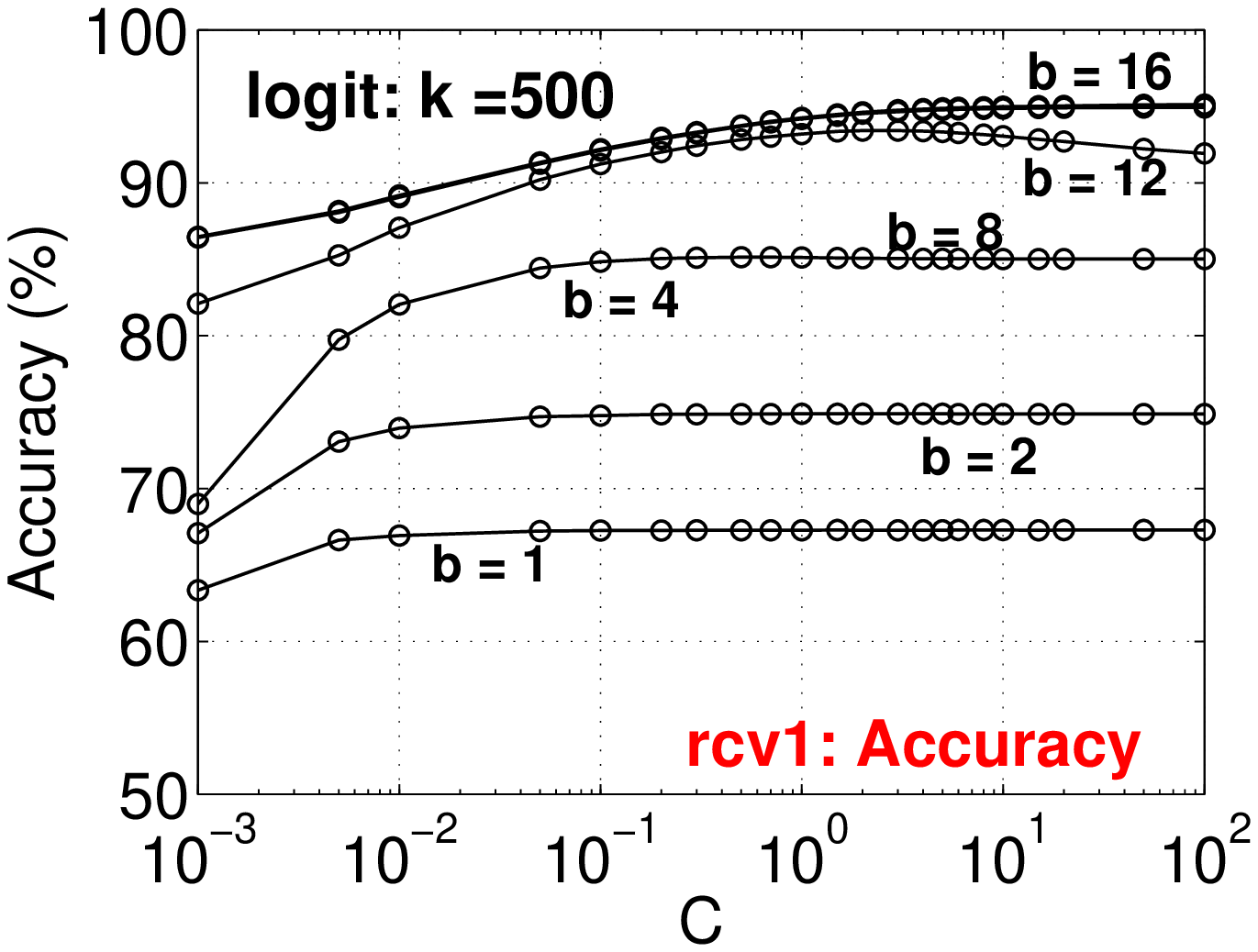}}

\vspace{-0.15in}

\caption{\textbf{Logistic regression test accuracy on rcv1}.  }\label{fig_rcv1_acc_logit}
\end{figure}

\begin{figure}[h!]
\mbox{
\includegraphics[width=1.7in]{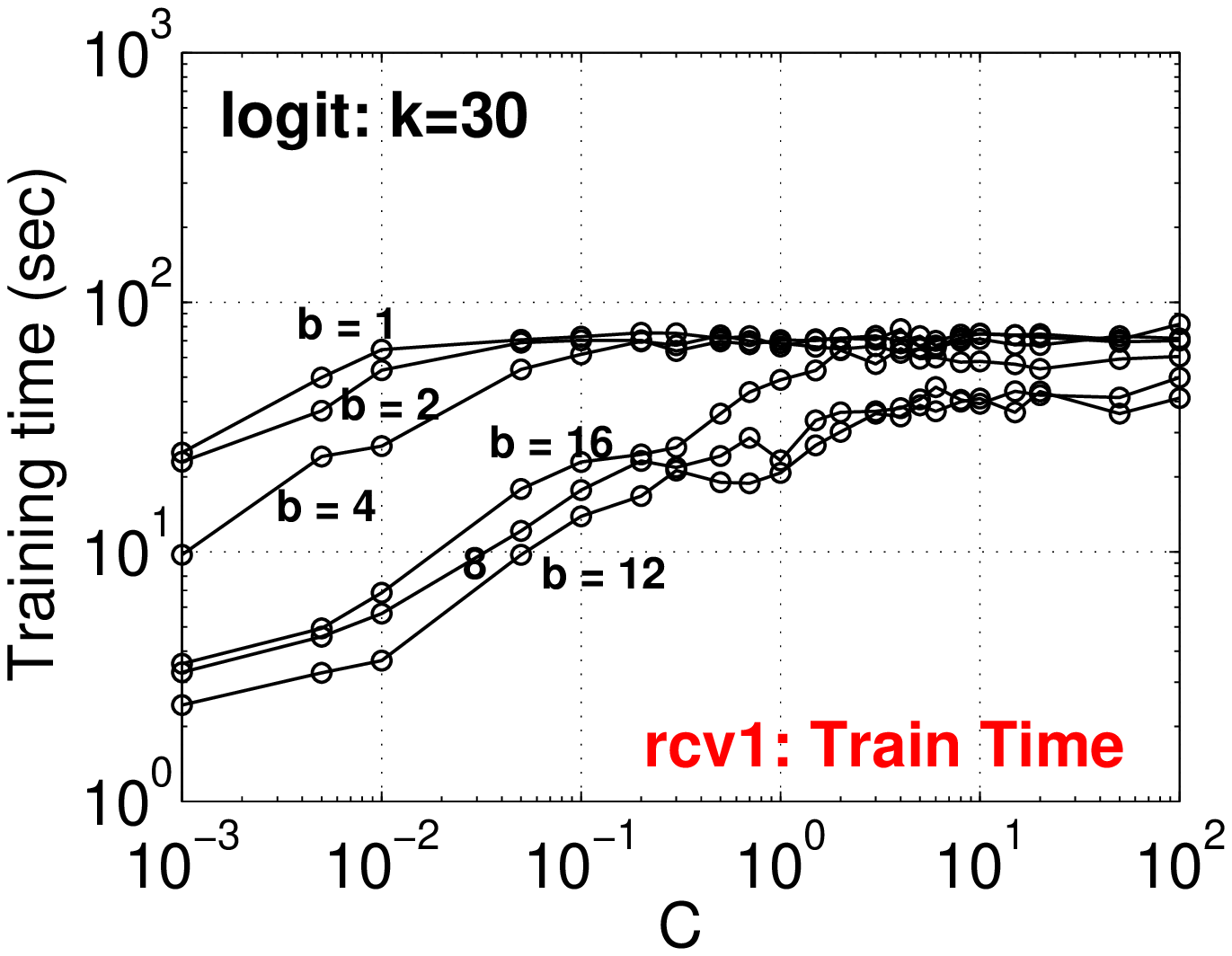}\hspace{-0.1in}
\includegraphics[width=1.7in]{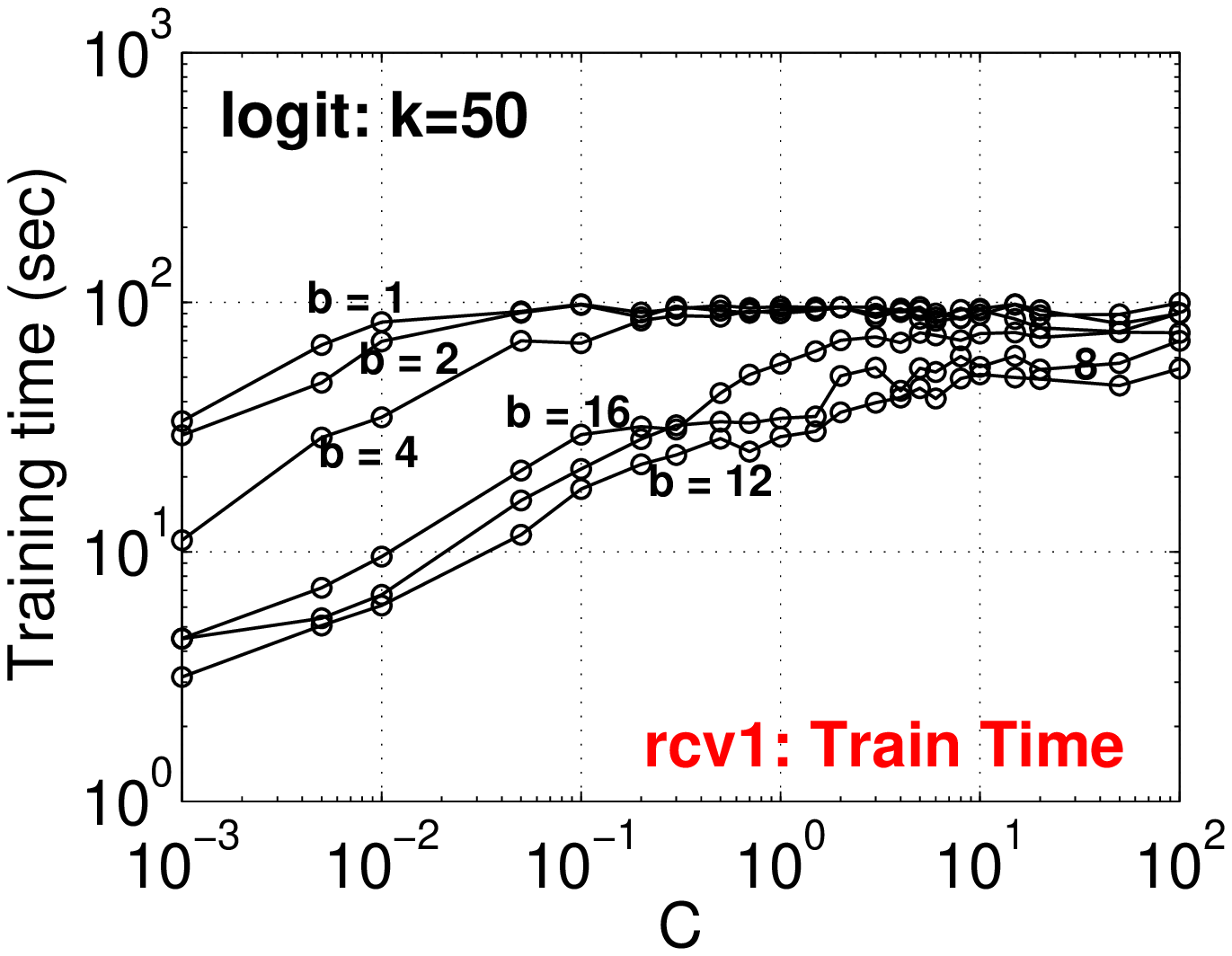}\hspace{-0.1in}
\includegraphics[width=1.7in]{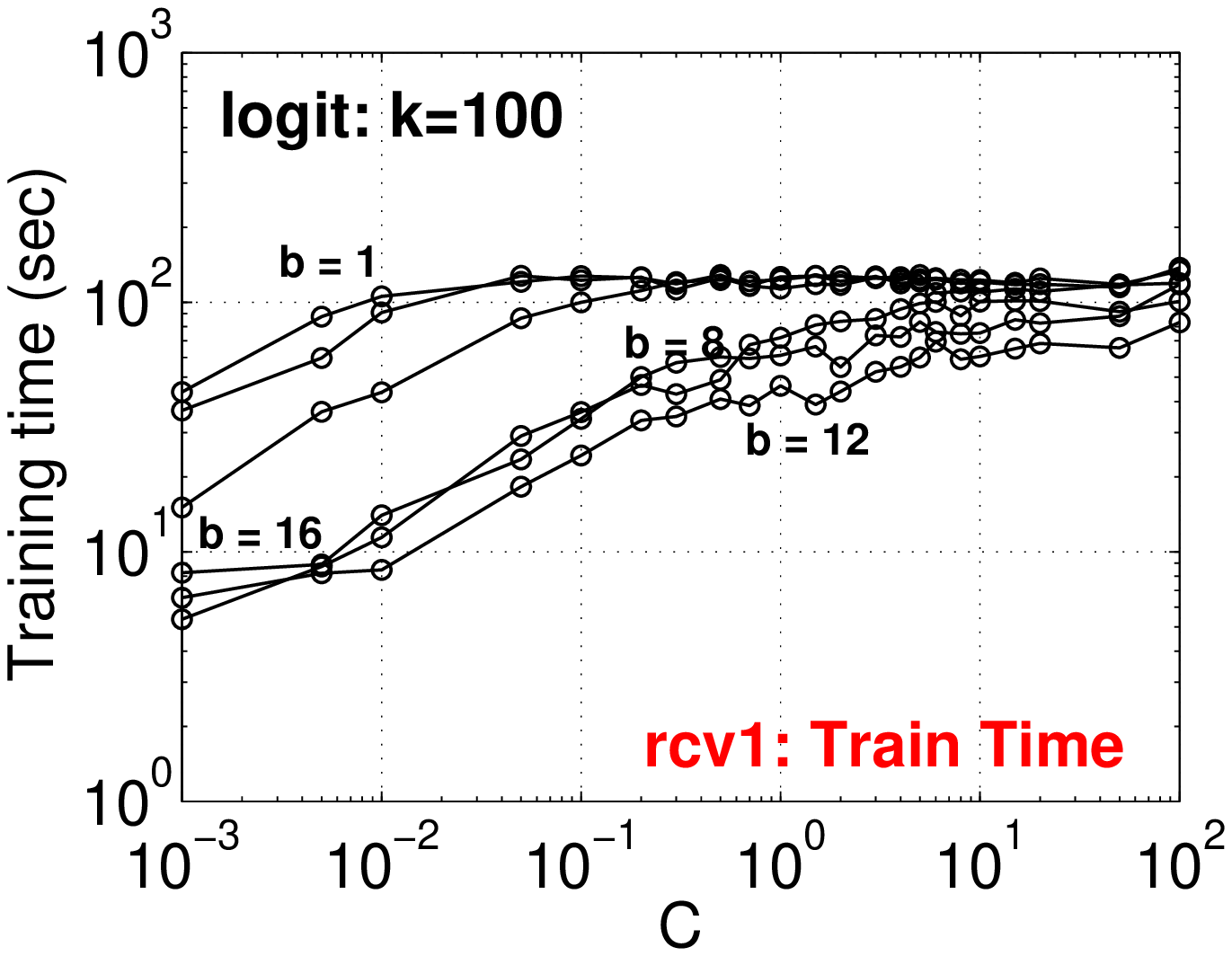}\hspace{-0.1in}
\includegraphics[width=1.7in]{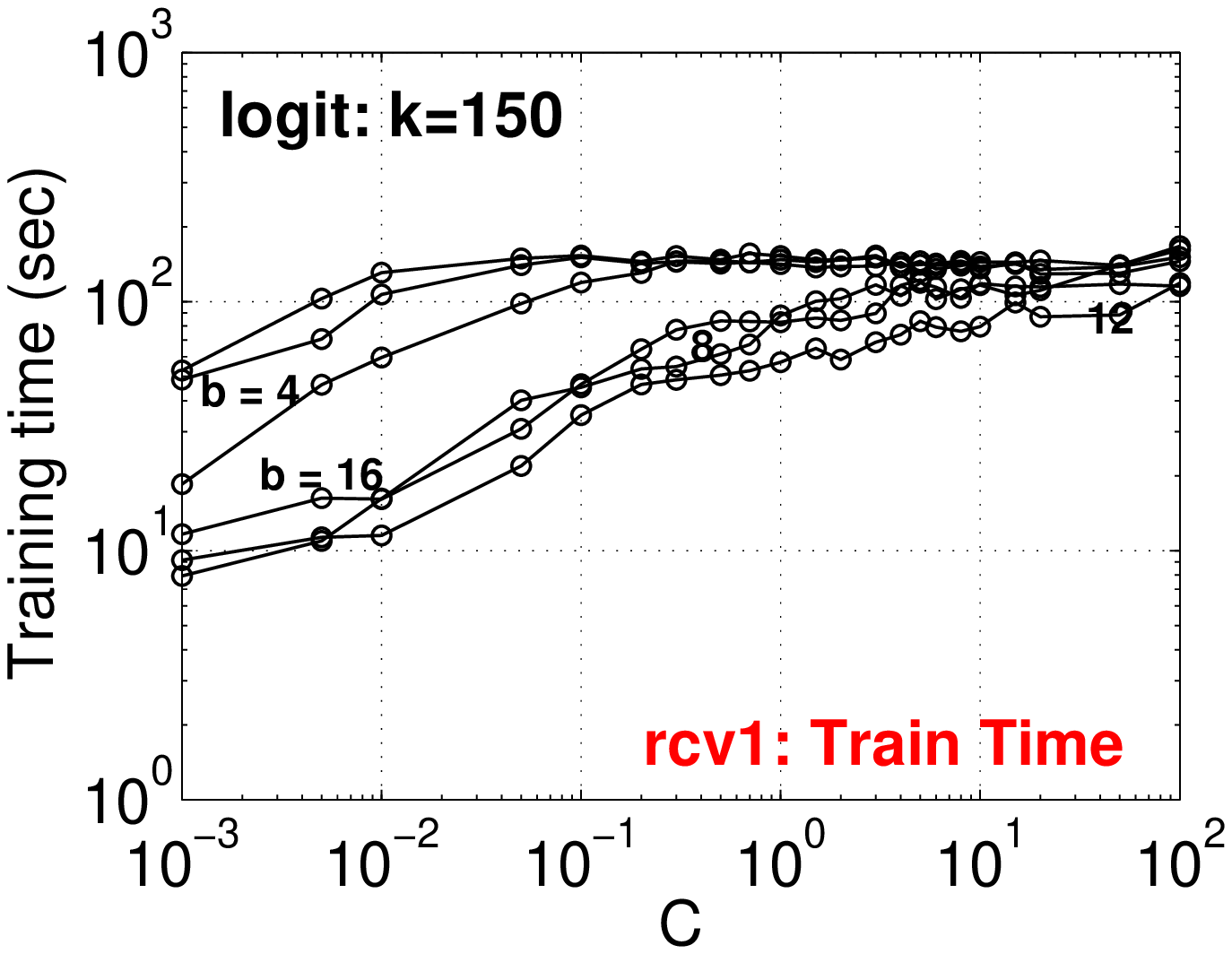}}

\mbox{
\includegraphics[width=1.7in]{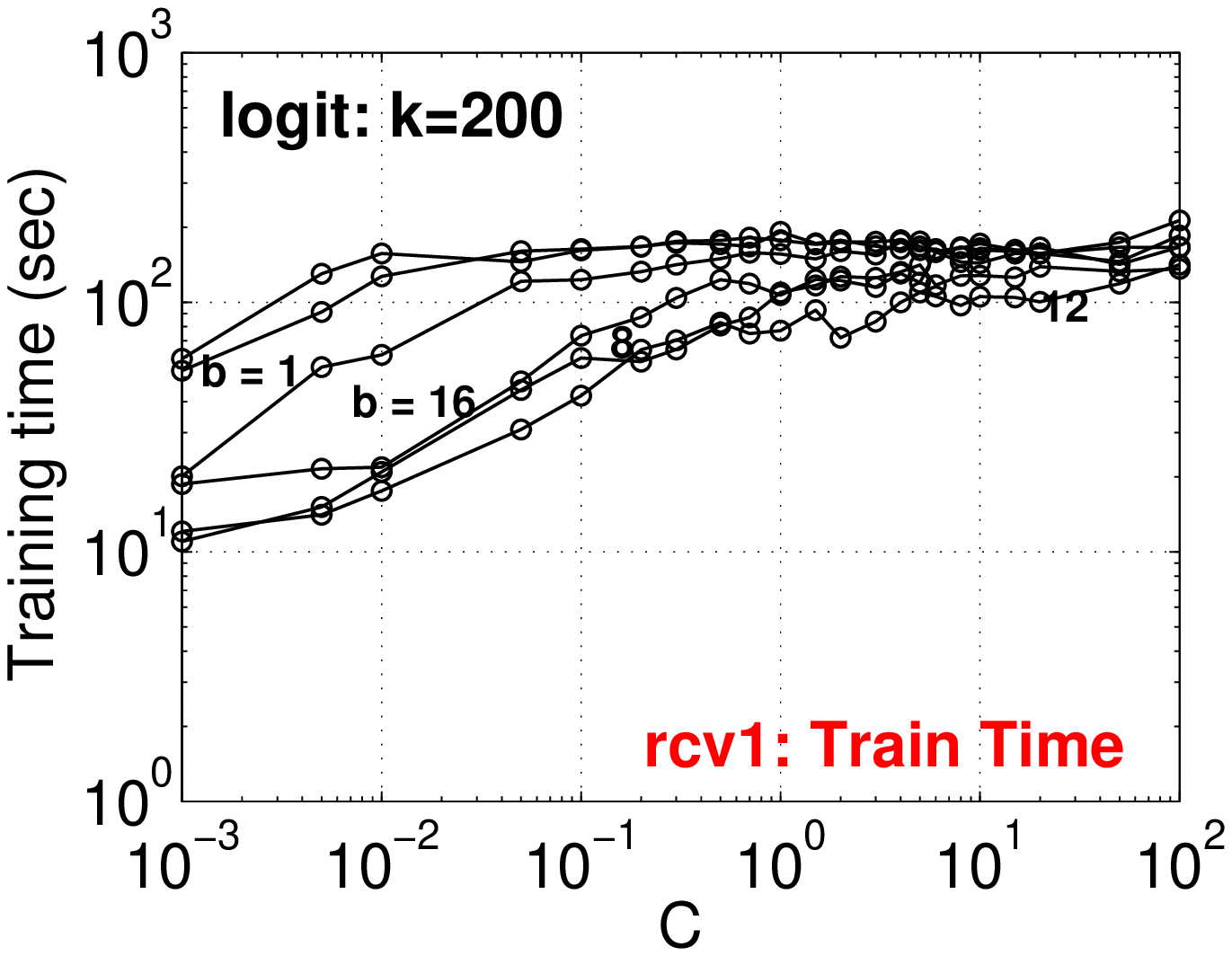}\hspace{-0.1in}
\includegraphics[width=1.7in]{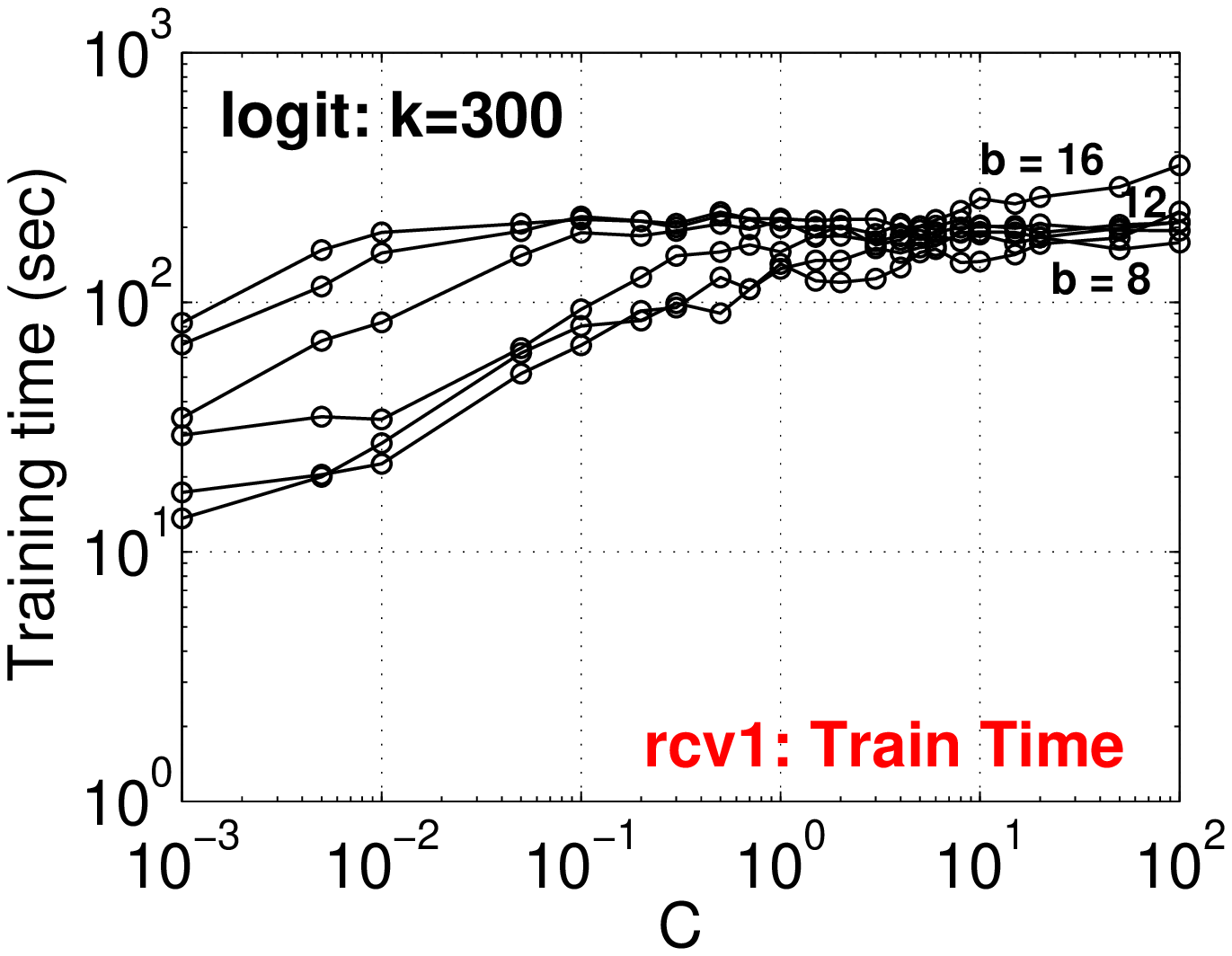}\hspace{-0.1in}
\includegraphics[width=1.7in]{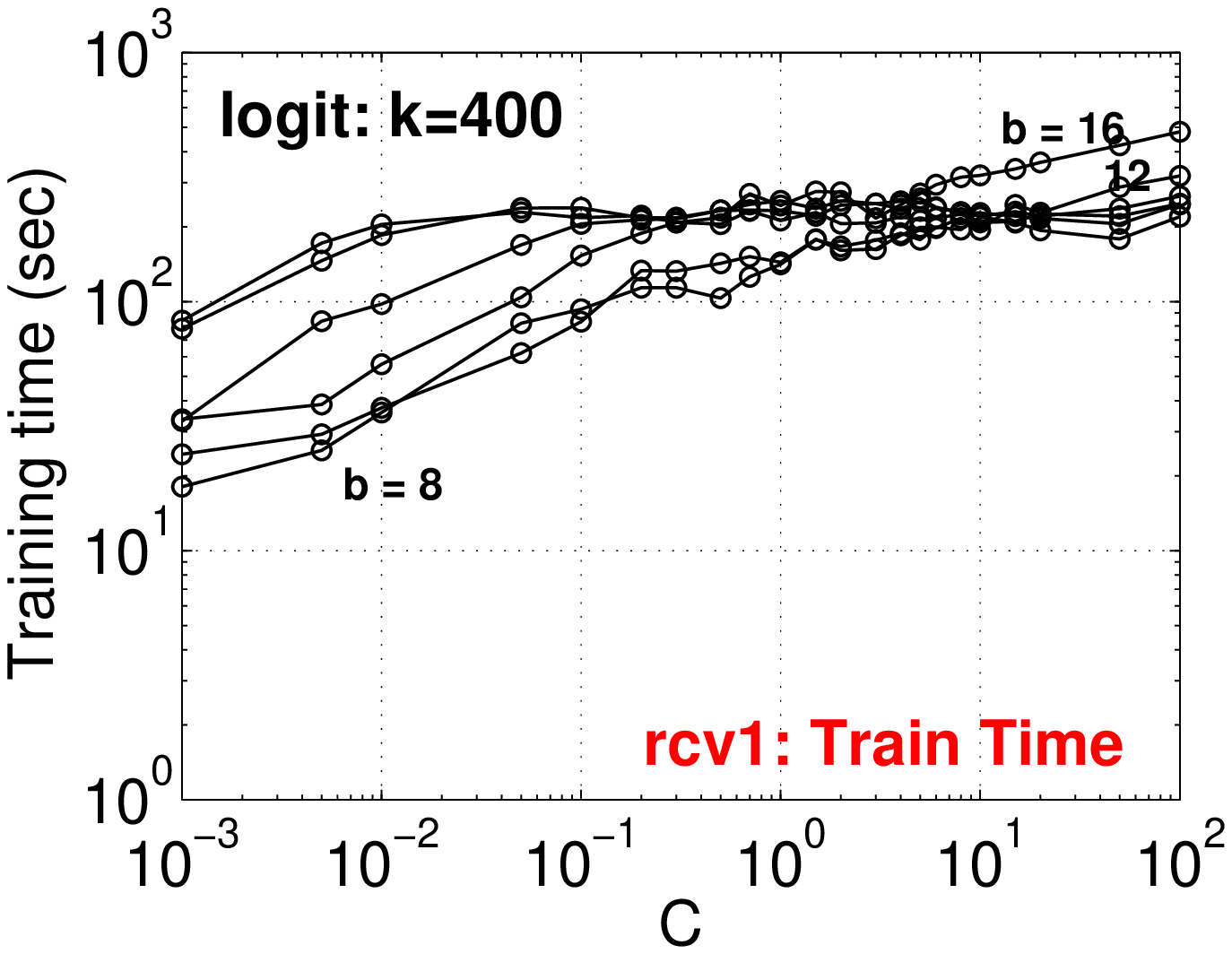}\hspace{-0.1in}
\includegraphics[width=1.7in]{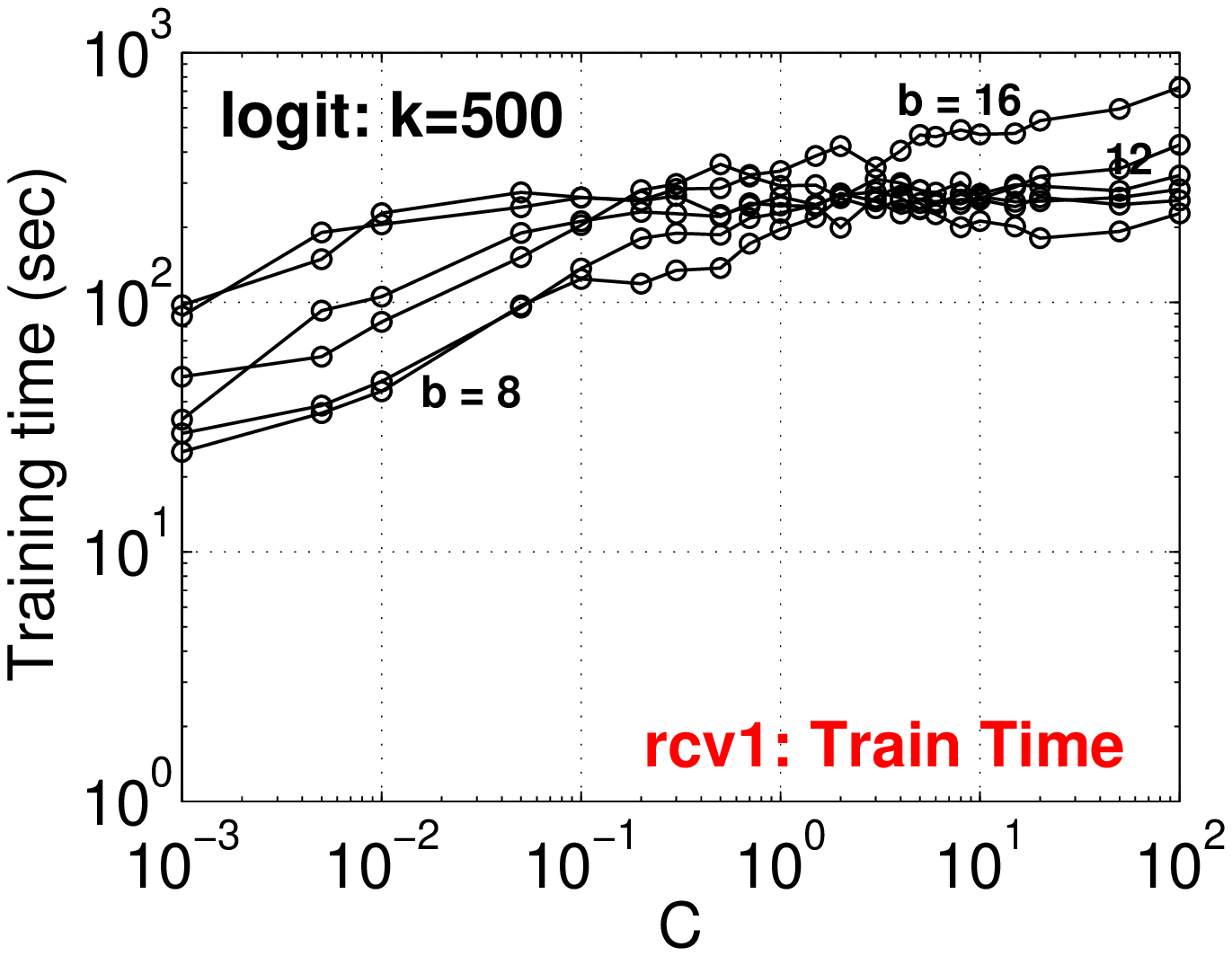}}

\vspace{-0.15in}

\caption{\textbf{Logistic regression training time on rcv1}.}\label{fig_rcv1_train_logit}
\end{figure}

Figure~\ref{fig_rcv1_acc_logit} presents the test accuracy and Figure~\ref{fig_rcv1_train_logit} presents the training time using logistic regression. Again, just like our experiments with SVM, using merely $k=30$ and $b=12$, we can achieve $>90\%$ test accuracies; and  using $k\geq 300$, we can achieve $>95\%$ test accuracies.

\clearpage

\section{Comparisons with Vowpal Wabbit (VW)}

The two methods, random projections~\cite{Article:Achlioptas_JCSS03, Proc:Li_Hastie_Church_KDD06} and Vowpal Wabbit (VW)~\cite{Proc:Weinberger_ICML2009,Article:Shi_JMLR09} are not limited to binary data (although for ultra high-dimensional used in the context of search, the data are often binary). The VW algorithm is also related to the Count-Min sketch~\cite{Article:Cormode_05}. In this paper, we use ``VW'' particularly for the hashing algorithm in~\cite{Proc:Weinberger_ICML2009}.

Since VW has the same variance as random projections (RP), we first provide a review for both RP and VW.

\subsection{Random Projections (RP)}

For convenience, we denote two $D$-dim data vectors by $u_1, u_2\in\mathbb{R}^{D}$. Again, the task is to estimate the inner product $a = \sum_{i=1}^D u_{1,i} u_{2,i}$.

The general idea is to multiply the data vectors, e.g., $u_1$ and $u_2$, by a random matrix $\{r_{ij}\}\in\mathbb{R}^{D\times k}$, where $r_{ij}$ is sampled i.i.d. from the following generic distribution with~\cite{Proc:Li_Hastie_Church_KDD06}
\begin{align}\label{eqn_r_ij}
E(r_{ij}) = 0, \ \ Var(r_{ij}) = 1,\ \ E(r_{ij}^3) = 0,\ \ E(r_{ij}^4) = s, \ \ s\geq 1.
\end{align}
We must have $s\geq 1$ because $Var(r_{ij}^2) = E(r_{ij}^4) - E^2(r_{ij}^2) = s-1 \geq 0$.

This generates two $k$-dim vectors, $v_1$ and $v_2$:
\begin{align}\notag
v_{1,j} = \sum_{i=1}^D u_{1,i}r_{ij},\hspace{0.2in} v_{2,j} = \sum_{i=1}^D u_{2,i}r_{ij}, \ \ \ j = 1, 2, ..., k
\end{align}

The general distributions which satisfy (\ref{eqn_r_ij}) include the standard normal distribution (in this case, $s=3$) and the ``sparse projection'' distribution specified as
\begin{align}\label{eqn_sparse_r}
r_{ij} = \sqrt{s}\times\left\{\begin{array}{ll} 1 & \text{with prob.}\ \frac{1}{2s} \\ 0 & \text{with prob.}\ 1-\frac{1}{s}\\ -1 & \text{with prob.}\ \frac{1}{2s} \end{array}\right.
\end{align}

\cite{Proc:Li_Hastie_Church_KDD06} provided the following unbiased estimator $\hat{a}_{rp,s}$ of $a$ and the general variance formula:
\begin{align}
&\hat{a}_{rp,s} = \frac{1}{k}\sum_{j=1}^k v_{1,j}v_{2,j}, \hspace{0.5in} E(\hat{a}_{rp,s}) = a = \sum_{i=1}^D u_{1,i} u_{2,i},\\\label{eqn_var_rp}
&Var(\hat{a}_{rp,s}) = \frac{1}{k}\left[\sum_{i=1}^D u_{1,i}^2\sum_{i=1}^D u_{2,i}^2 + \left(\sum_{i=1}^D u_{1,i}u_{2,i}\right)^2 +(s-3)\sum_{i=1}^D u_{1,i}^2u_{2,i}^2\right]
\end{align}
which means $s=1$ achieves the smallest  variance. The only elementary distribution we know that satisfies (\ref{eqn_r_ij}) with $s=1$ is the two point distribution in $\{-1, 1\}$ with equal probabilities, i.e., (\ref{eqn_sparse_r}) with $s=1$.

\subsection{Vowpal Wabbit (VW)}

Again, in this paper, ``VW'' always refers to the particular algorithm in~\cite{Proc:Weinberger_ICML2009}. VW may be viewed as a ``bias-corrected'' version of the  Count-Min (CM) sketch algorithm~\cite{Article:Cormode_05}. In the original CM  algorithm,  the key step  is to independently and uniformly hash elements of the data vectors to buckets $\in\{1, 2, 3, ..., k\}$ and the hashed value is the sum of the elements in the bucket. That is $h(i) = j$ with probability $\frac{1}{k}$, where $j \in\{1, 2, ..., k\}$. For convenience, we introduce an indicator function:
\begin{align}\notag
I_{ij} =\left\{\begin{array}{ll}
1 &\text{if } h(i) = j\\
0 &\text{otherwise}
\end{array}
\right.\end{align}
which allow  us to write the hashed data as
\begin{align}\notag
w_{1,j} = \sum_{i=1}^D u_{1,i}I_{ij},\hspace{0.5in} w_{2,j} = \sum_{i=1}^D u_{2,i}I_{ij}
\end{align}

The estimate $\hat{a}_{cm} = \sum_{j=1}^k w_{1,j} w_{2,j}$ is (severely) biased for the task of estimating the inner products. \cite{Proc:Weinberger_ICML2009} proposed a creative  method for bias-correction, which consists of pre-multiplying (element-wise) the original data vectors with a random vector whose entries are sampled i.i.d. from the  two-point distribution in $\{-1,1\}$ with equal probabilities, which corresponds to $s =1$ in (\ref{eqn_sparse_r}).

\cite{Report:HashLearning11} considered a more general situation, for any  $s\geq 1$. After applying multiplication and hashing on $u_1$ and $u_2$ as in \cite{Proc:Weinberger_ICML2009}, the resultant vectors $g_1$ and $g_2$ are
{\begin{align}
g_{1,j} = \sum_{i=1}^D u_{1,i} r_i I_{ij}, \hspace{0.5in} g_{2,j} = \sum_{i=1}^D u_{2,i} r_i I_{ij}, \ \ \ j = 1, 2, ..., k
\end{align}}
where $r_i$ is defined as in (\ref{eqn_r_ij}), i.e.,  $E(r_i) = 0, \ E(r_i^2) = 1, \ E(r_i^3) = 0, \ E(r_i^4) = s$. \cite{Report:HashLearning11} proved that
\begin{align}
&\hat{a}_{vw,s} = \sum_{j=1}^k g_{1,j} g_{2,j},\hspace{0.2in}  E(\hat{a}_{vw,s}) = \sum_{i=1}^D u_{1,i}u_{2,i} =a, \\\label{eqn_var_vw}
&Var(\hat{a}_{vw,s}) = (s-1)\sum_{i=1}^Du_{1,i}^2u_{2,i}^2 + \frac{1}{k}\left[\sum_{i=1}^D u_{1,i}^2\sum_{i=1}^D u_{2,i}^2 + \left(\sum_{i=1}^D u_{1,i}u_{2,i}\right)^2-2\sum_{i=1}^D u_{1,i}^2u_{2,i}^2\right]
\end{align}

The variance (\ref{eqn_var_vw}) says we do need $s=1$, otherwise the additional term $(s-1)\sum_{i=1}^Du_{1,i}^2u_{2,i}^2$ will not vanish even as the sample size $k\rightarrow\infty$. In other words, the choice of random distribution in VW is essentially the only option if we want to remove the bias by pre-multiplying the data vectors (element-wise) with a vector of random variables. Of course, once we let $s=1$, the variance (\ref{eqn_var_vw}) becomes identical to the variance of random projections (\ref{eqn_var_rp}).

\subsection{Comparing $b$-bit Minwise Hashing with RP and VW in Terms of Variances}

Each sample of $b$-bit minwise hashing requires exactly $b$ bits of storage. For VW, if we consider the number of bins $k$ is smaller than the number of nonzeros in each data vector, then the resultant hashed data vectors are dense and we probably need 32 bits or 16 bits per hashed data entry (sample). \cite{Report:HashLearning11} demonstrated that if each sample of VW needs 32 bits, then VW needs $10\sim 100$ (or even 10000) times more space than $b$-bit minwise hashing in order to achieve the same variance. Of course, when $k$ is much larger than the number of nonzeros in each data vector, then the resultant hashed data vector will be sparse and the storage would be similar to the original data size.

One reason why VW is not accurate is because the variance (\ref{eqn_var_vw}) (for $s=1$) is dominated by the product of two marginal squared $l_2$ norms $\sum_{i=1}^D u_{1,i}^2\sum_{i=1}^D u_{2,i}^2$ even when the inner product is zero.

\subsection{Experiments}

We experiment with VW using $k = 2^5, 2^6, 2^7, 2^8, 2^9, 2^{10}, 2^{11}, 2^{12}, 2^{13}$, and $2^{14}$. Note that $2^{14} = 16384$. It is difficult to train LIBLINEAR with $k=2^{15}$ because the training size of the hashed  data by VW is  close to 48 GB when $k=2^{15}$.

\begin{figure}[h!]
\begin{center}
\mbox{
\includegraphics[width=1.7in]{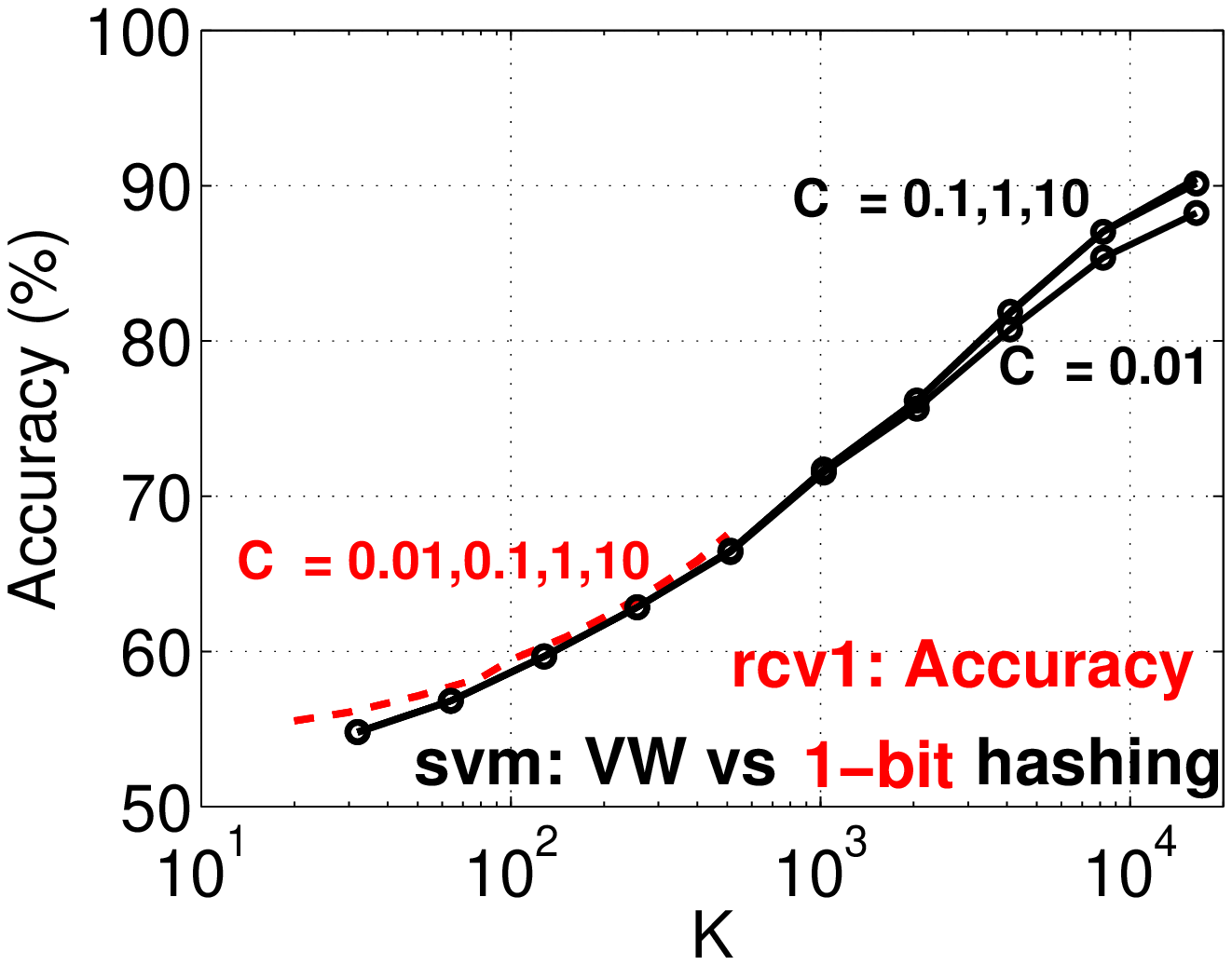}\hspace{0.2in}
\includegraphics[width=1.7in]{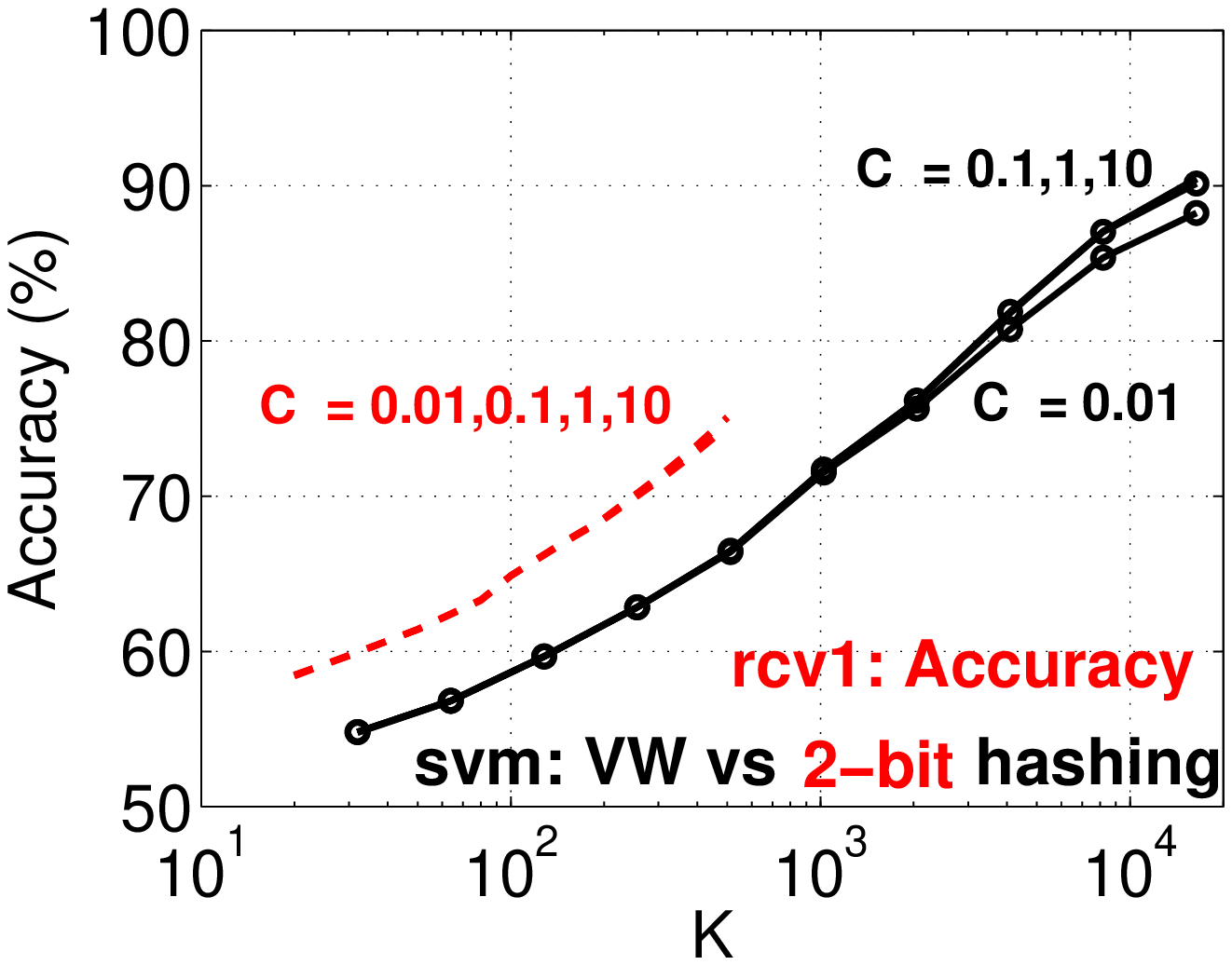}\hspace{0.2in}
\includegraphics[width=1.7in]{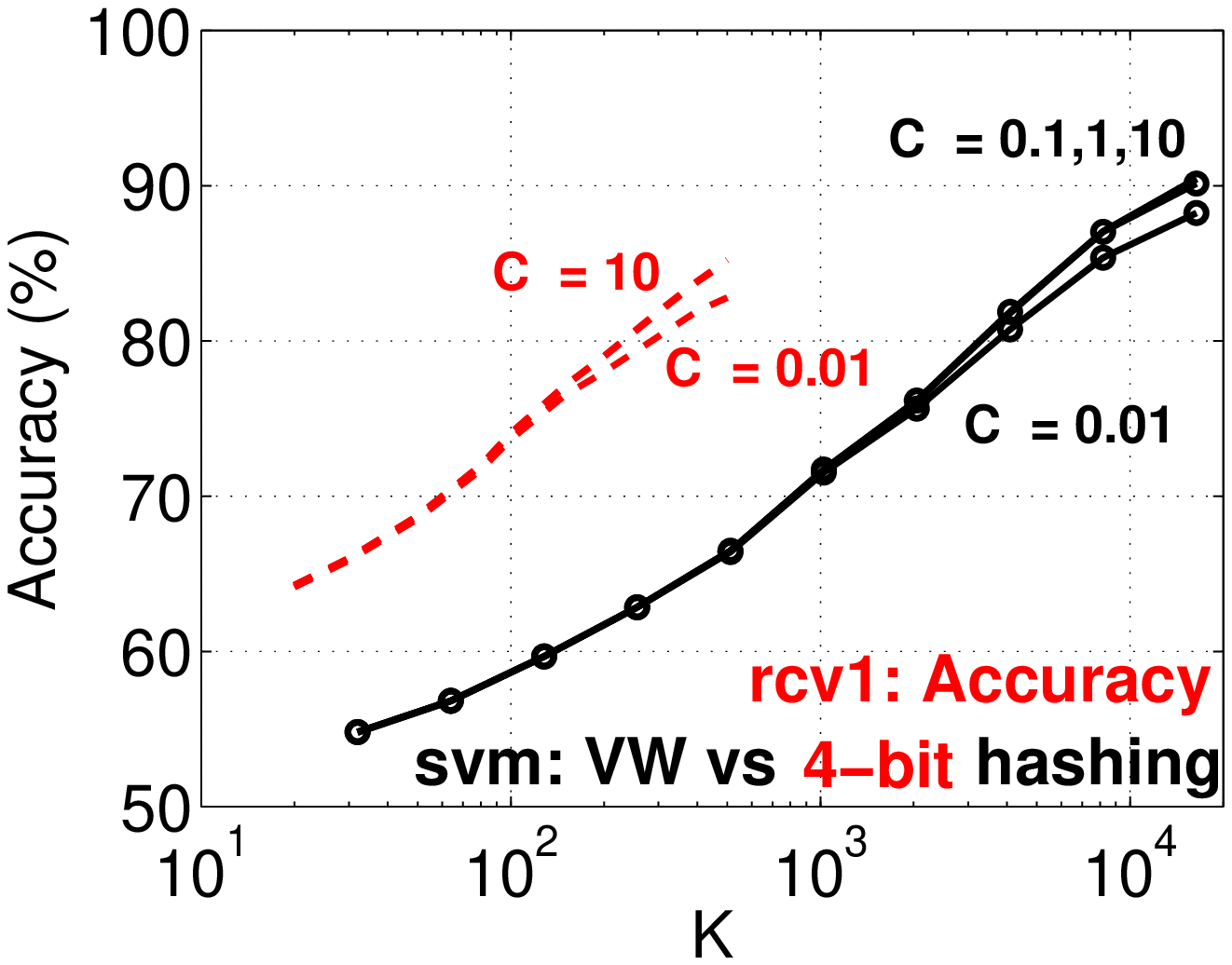}}

\mbox{
\includegraphics[width=1.7in]{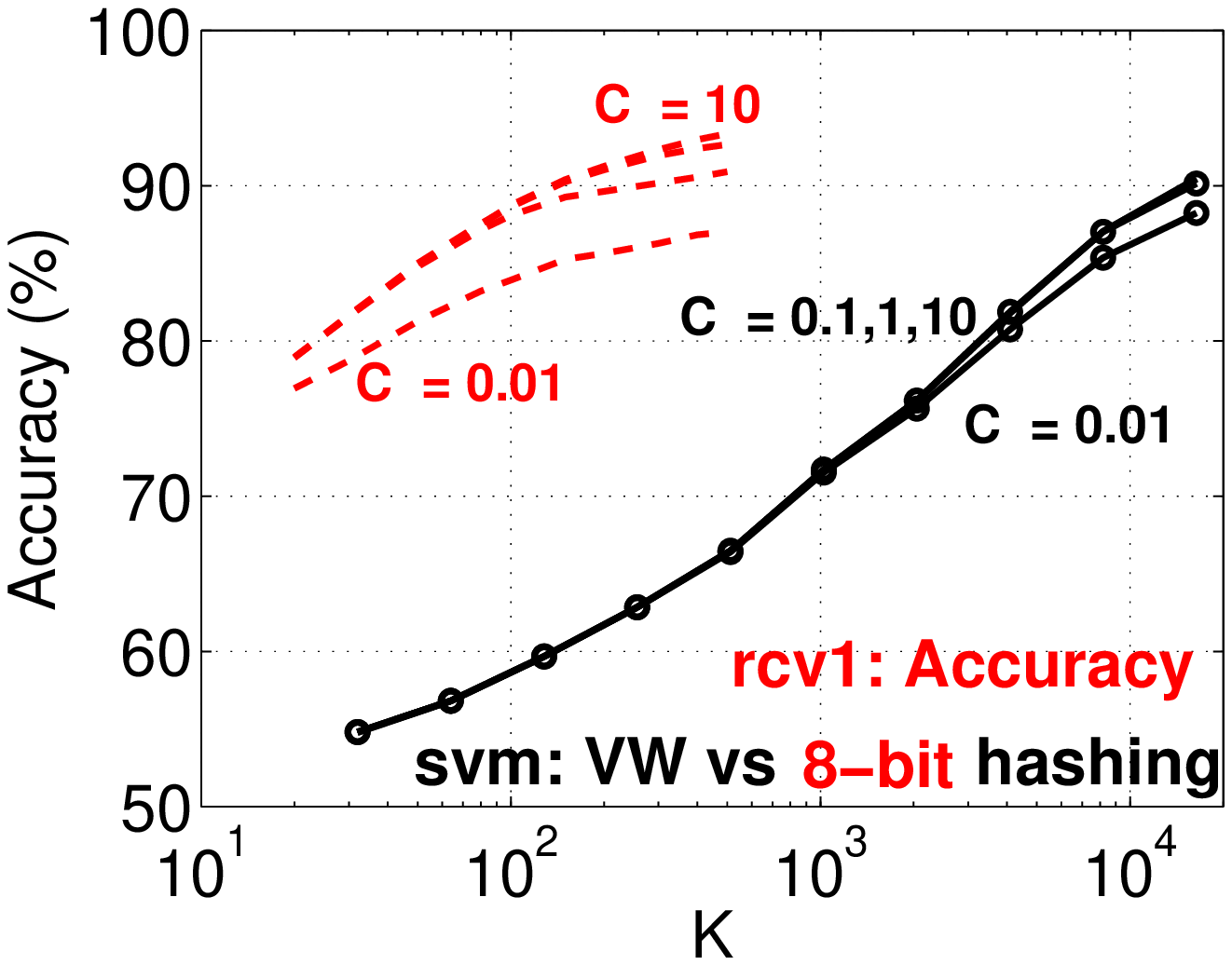}\hspace{0.2in}
\includegraphics[width=1.7in]{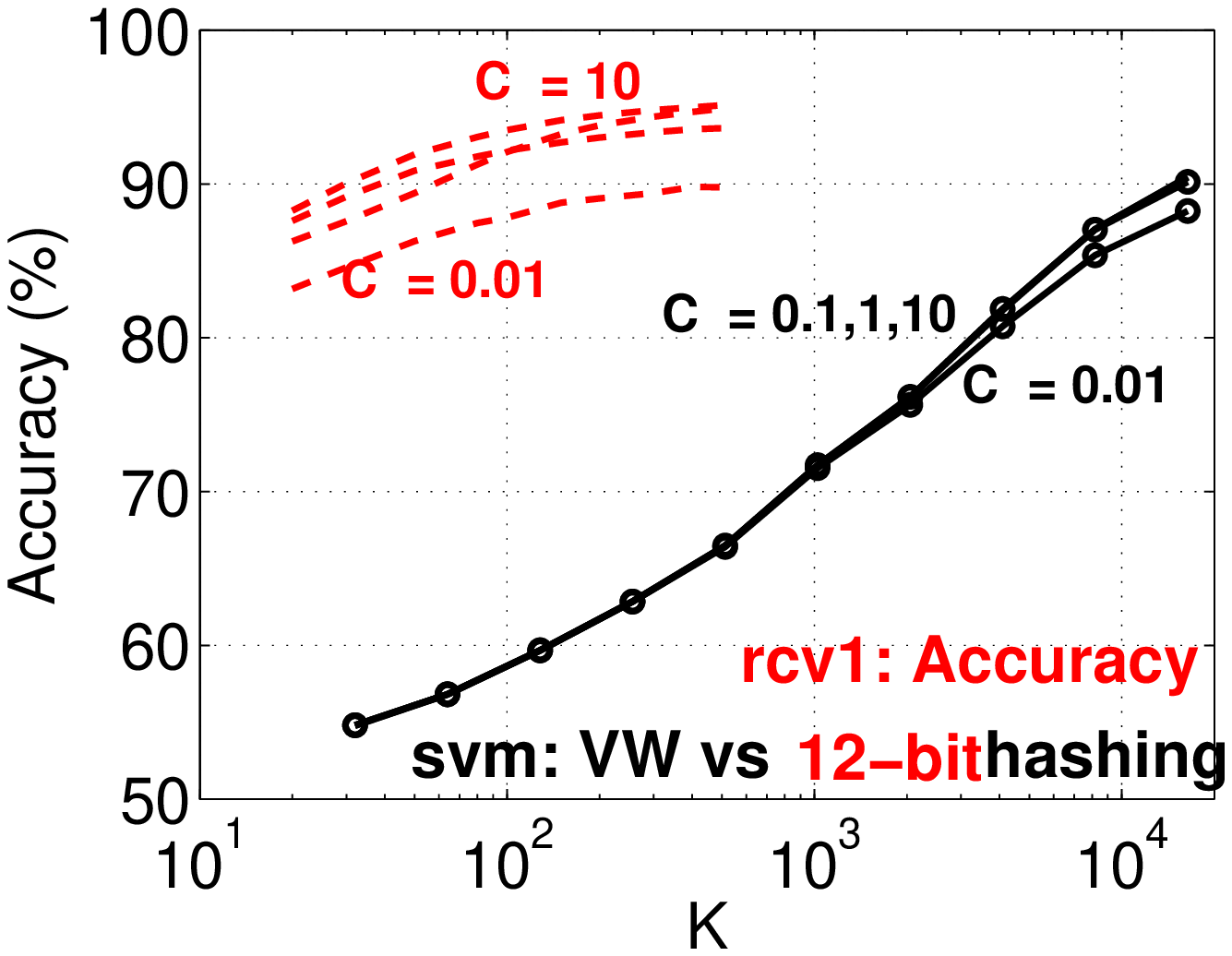}\hspace{0.2in}
\includegraphics[width=1.7in]{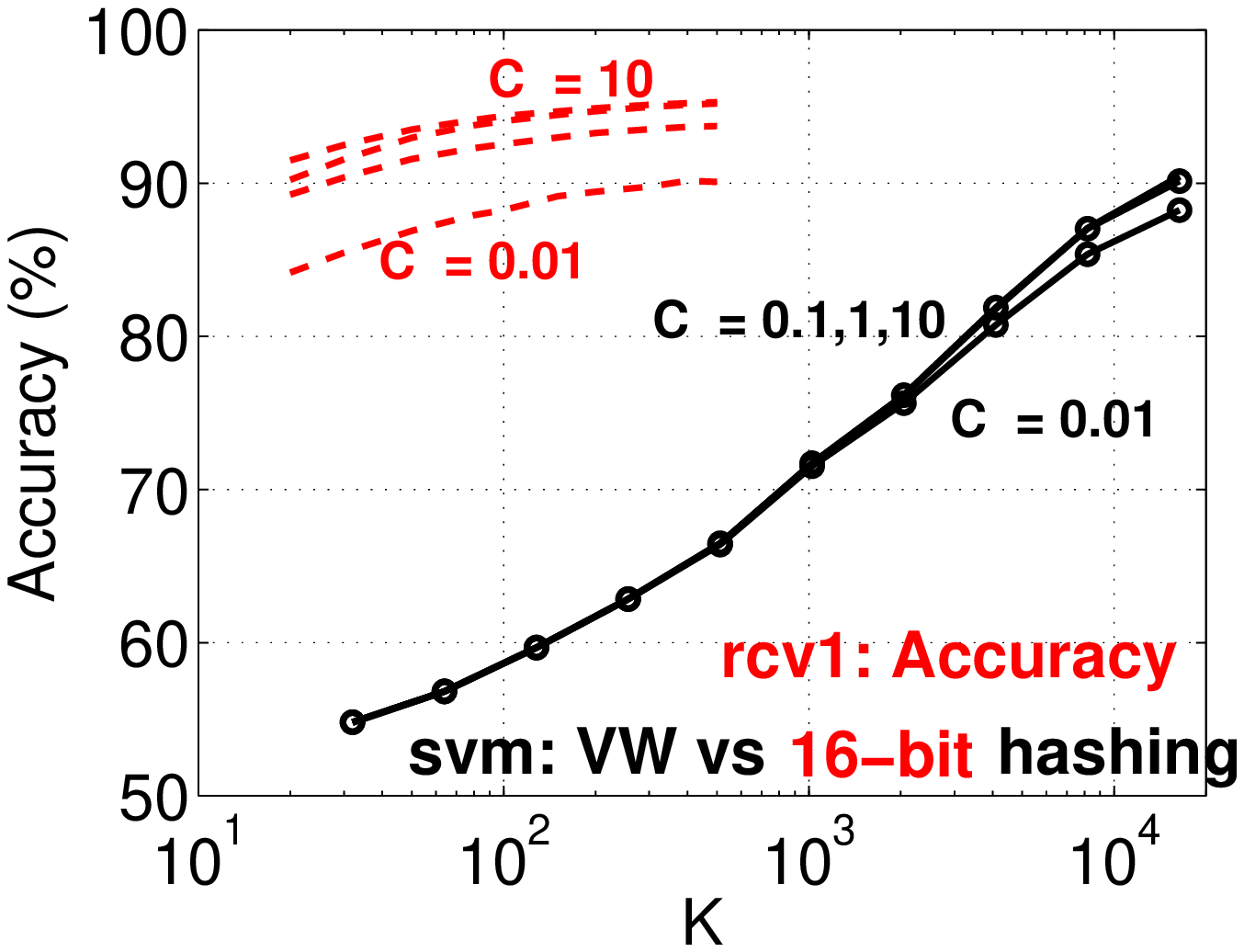}}

\end{center}
\vspace{-0.25in}
\caption{\textbf{SVM test accuracy on rcv1} for comparing VW (solid) with $b$-bit minwise hashing (dashed). Each panel plots the same results for VW and results for $b$-bit minwise hashing for a different $b$. We select $C = 0.01, 0.1, 1, 10$. }\label{fig_rcv1_acc_vw}
\end{figure}

\begin{figure}[h!]

\begin{center}
\mbox{
\includegraphics[width=1.7in]{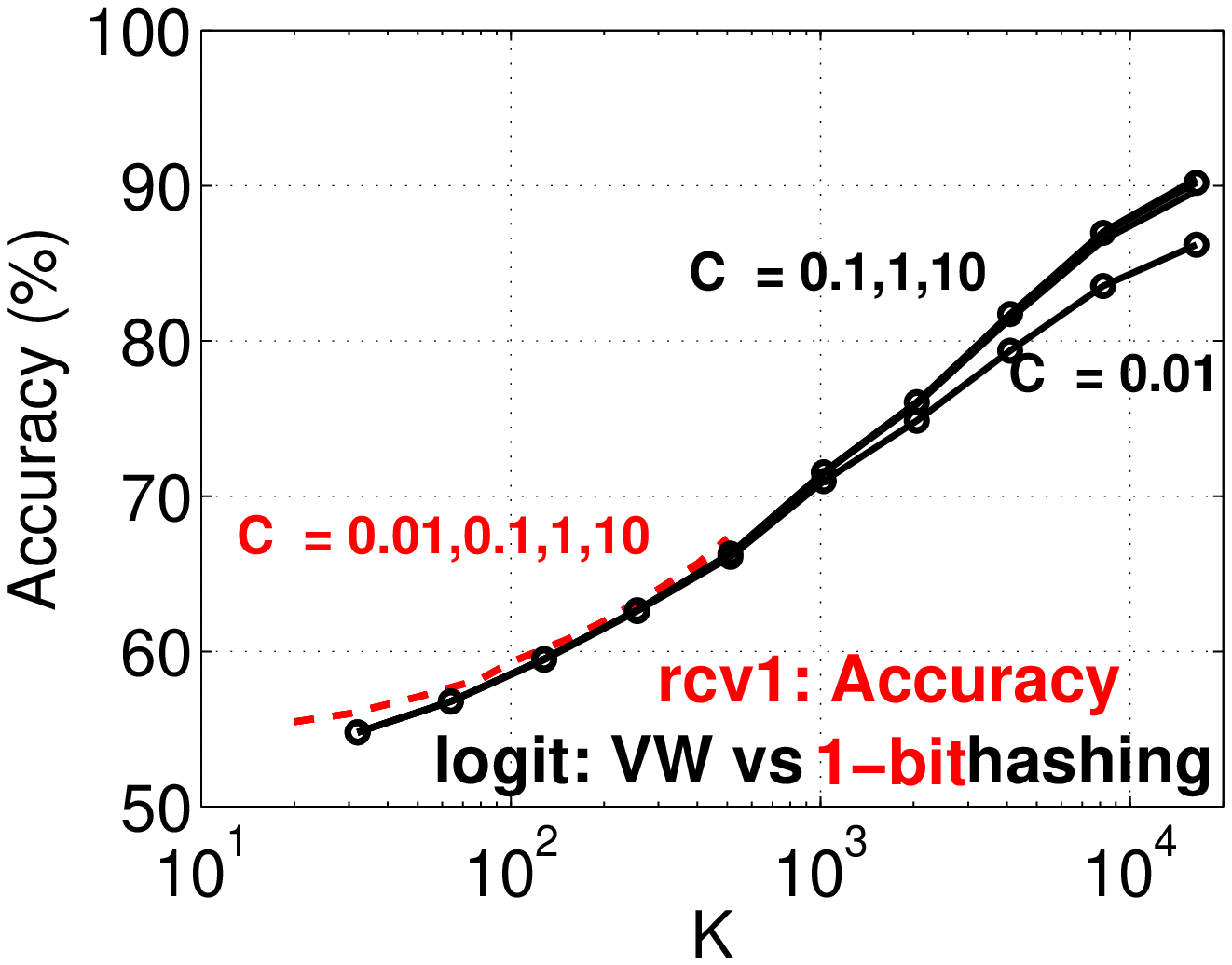}\hspace{0.2in}
\includegraphics[width=1.7in]{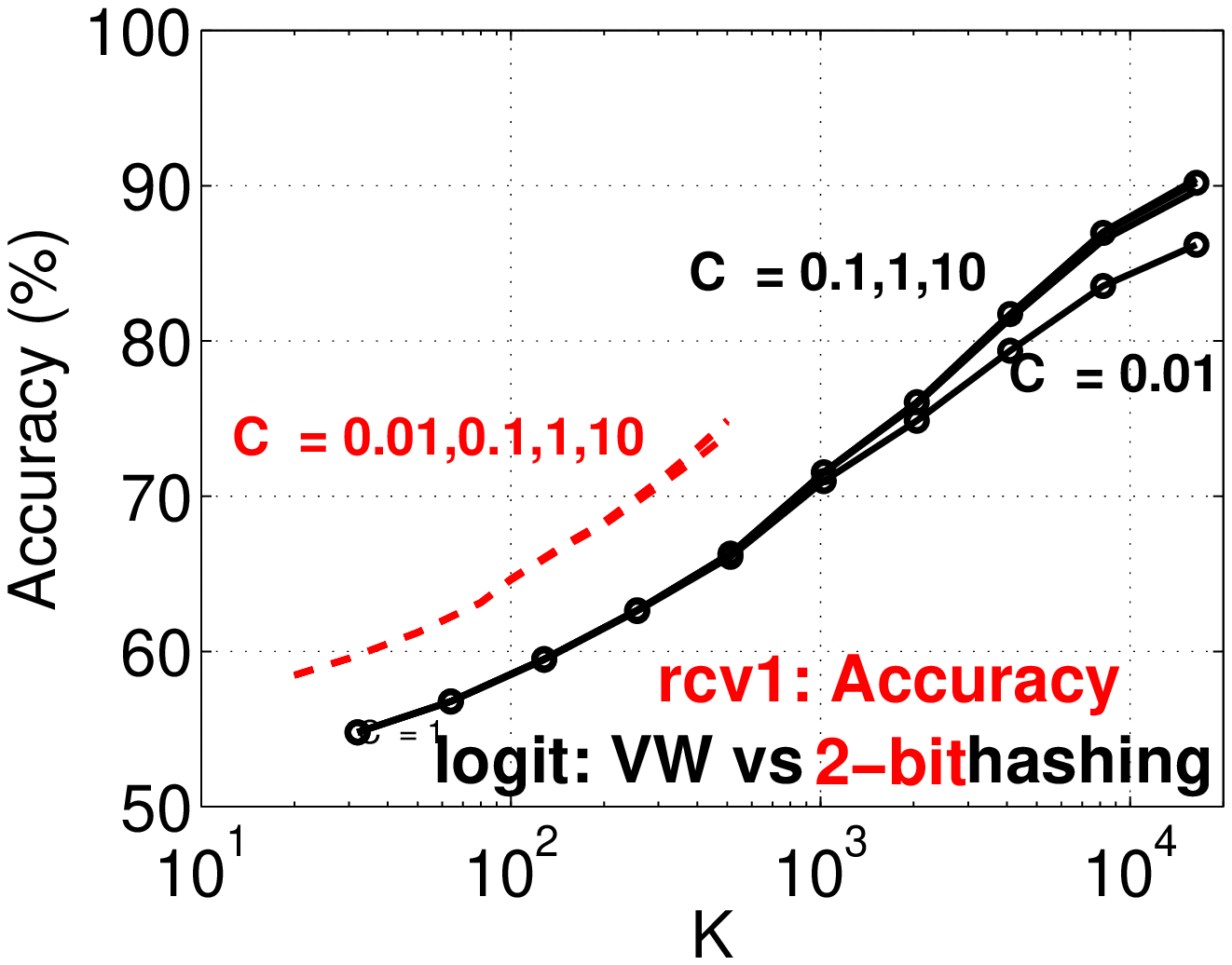}\hspace{0.2in}
\includegraphics[width=1.7in]{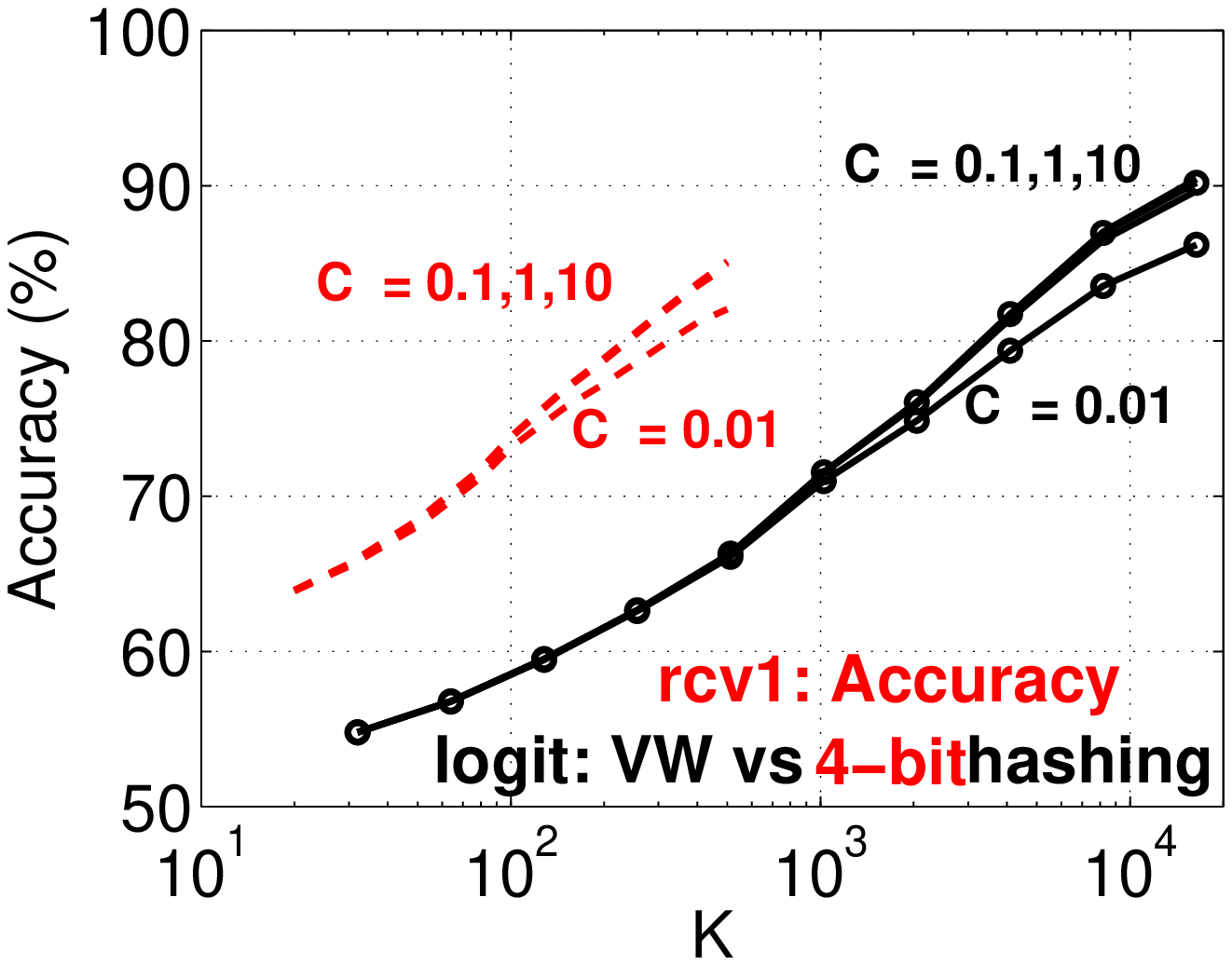}}

\mbox{
\includegraphics[width=1.7in]{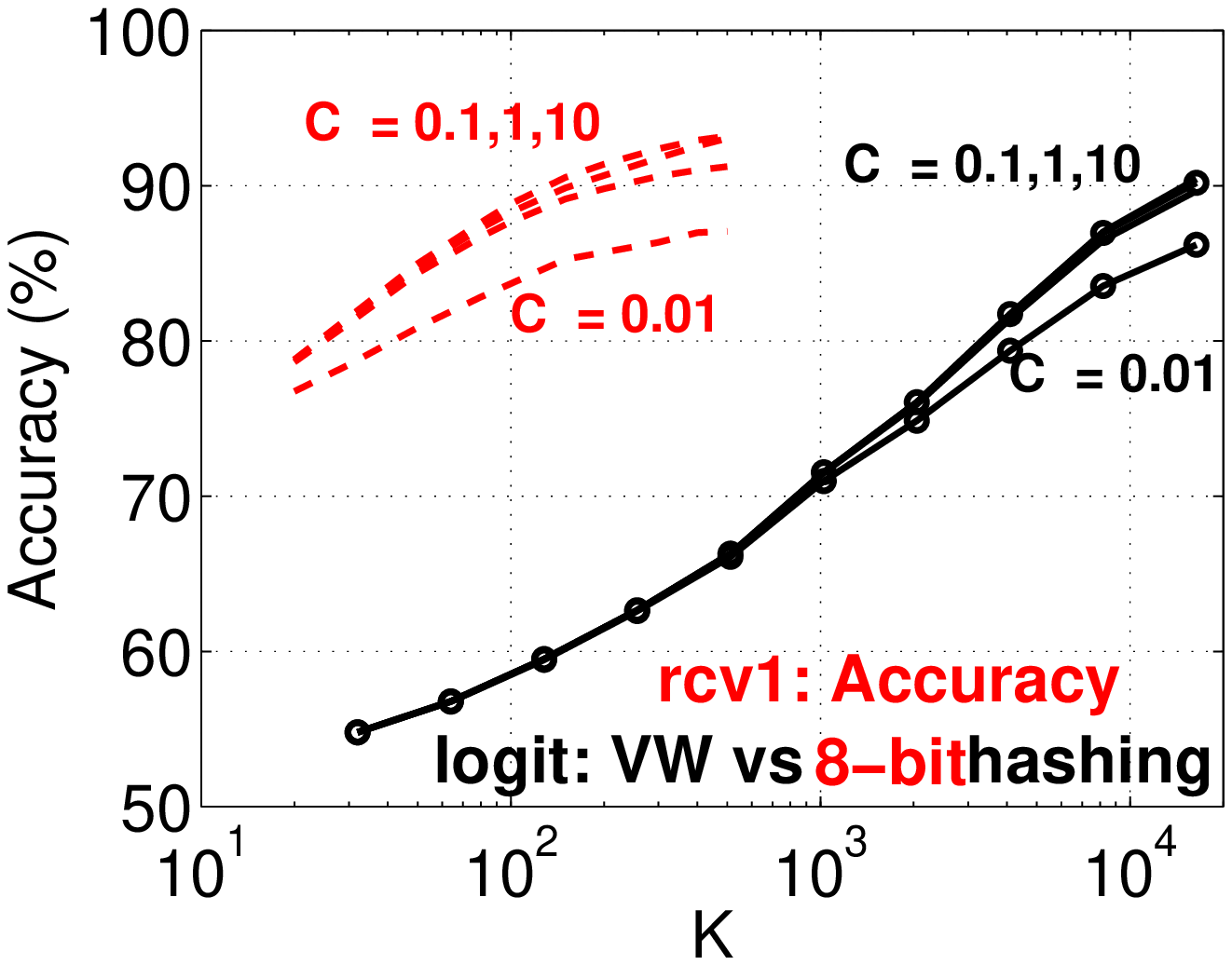}\hspace{0.2in}
\includegraphics[width=1.7in]{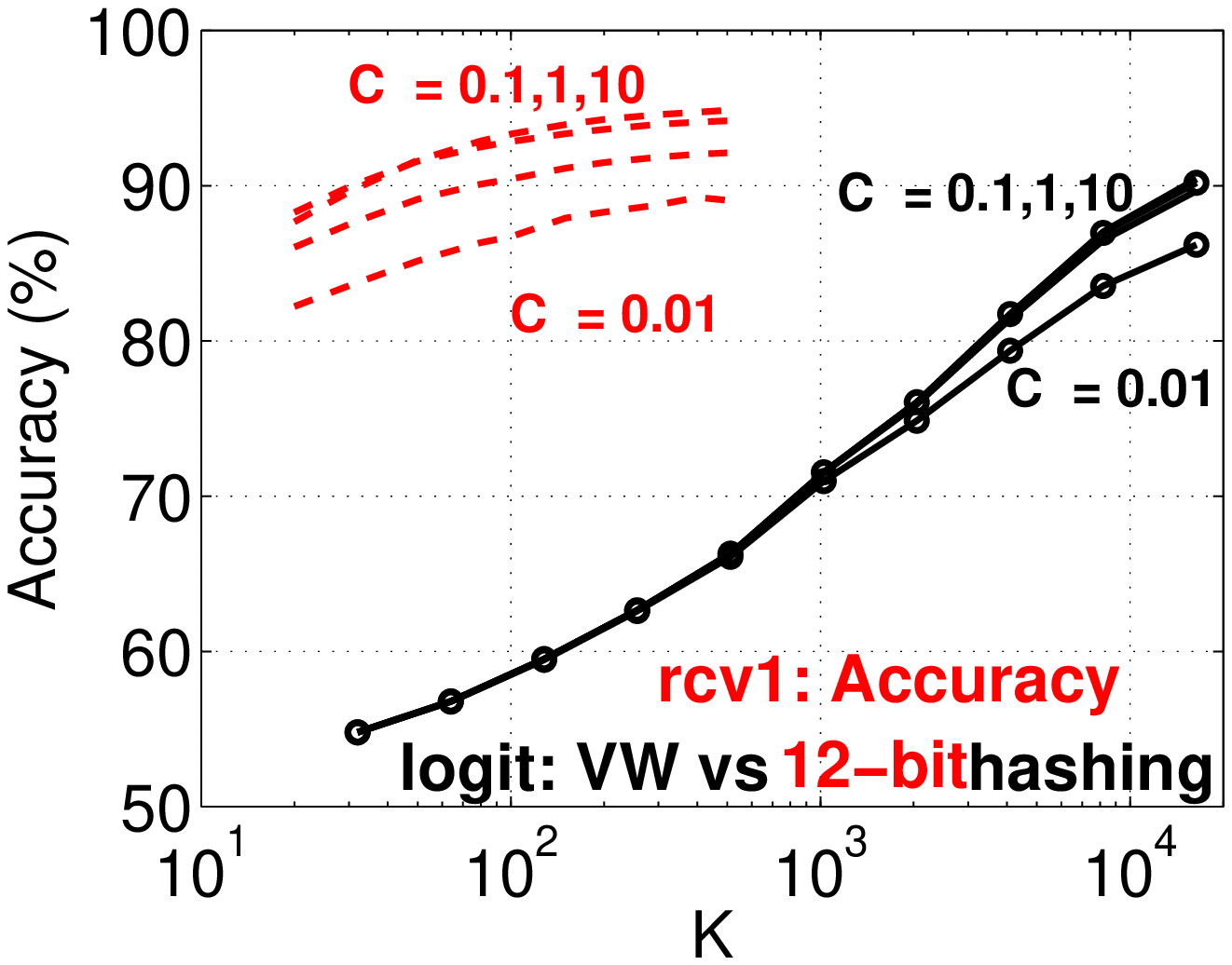}\hspace{0.2in}
\includegraphics[width=1.7in]{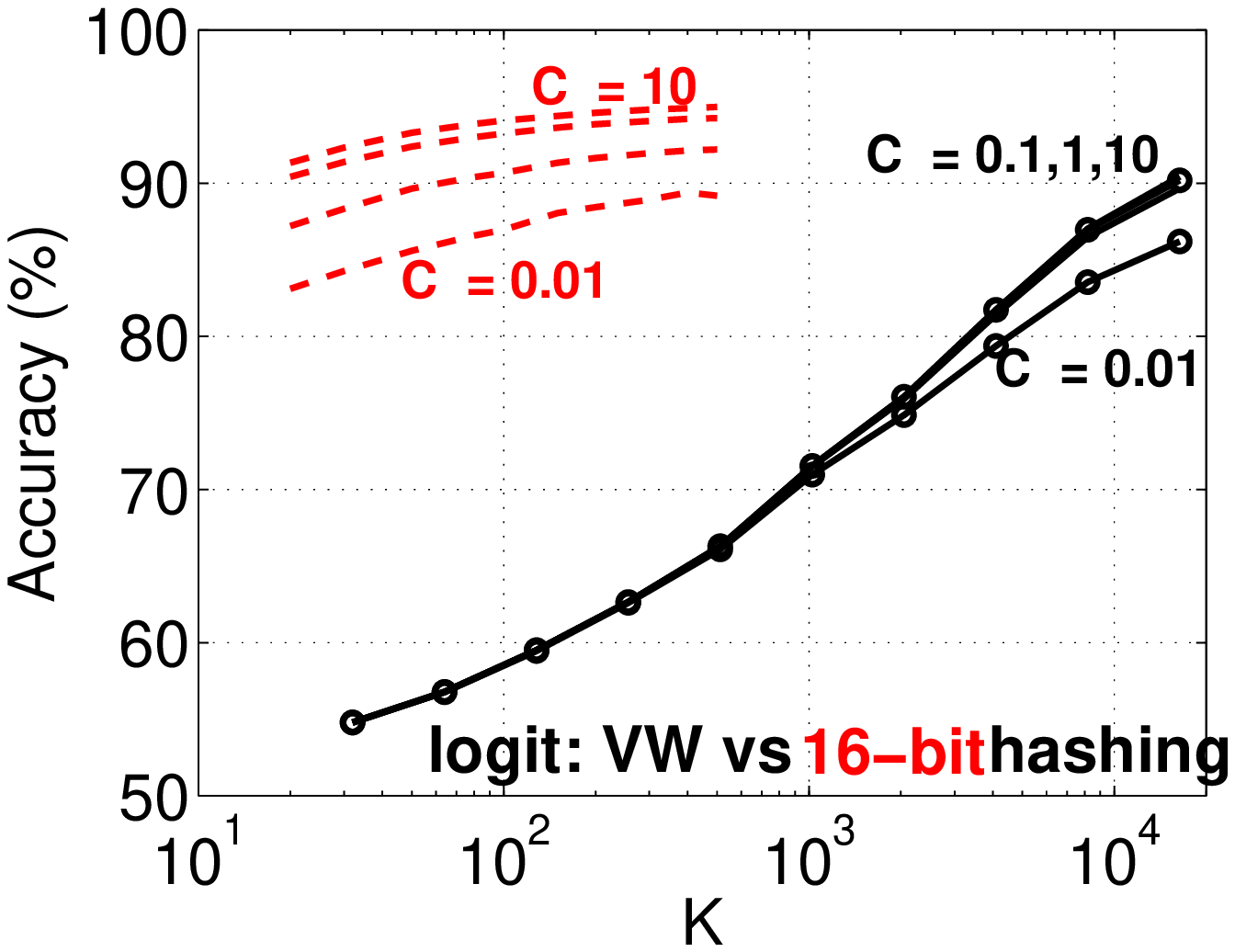}}

\end{center}

\vspace{-0.25in}

\caption{\textbf{Logistic Regression test accuracy on rcv1} for comparing VW with $b$-bit minwise hashing. }\label{fig_rcv1_acc_vw_logit}
\end{figure}

Figure~\ref{fig_rcv1_acc_vw} and Figure~\ref{fig_rcv1_acc_vw_logit}  plot the test accuracies for SVM and logistic regression, respectively. In each figure, every panel has the same set of solid curves for VW but a different set of dashed curves for different $b$-bit minwise hashing. Since $k$ ranges very large, here we choose to present the test accuracies against $k$. Representative $C$ values (0.01, 0.1, 1, 10) are selected for the presentations.

From Figures~\ref{fig_rcv1_acc_vw} and~\ref{fig_rcv1_acc_vw_logit}, we can see clearly that $b$-bit minwise hashing is substantially more accurate than VW at the same storage. In other words, in order to achieve the same accuracy, VW will require substantially more storage than $b$-bit minwise hashing. 

\begin{figure}[h!]

\begin{center}
\mbox{
\includegraphics[width=1.7in]{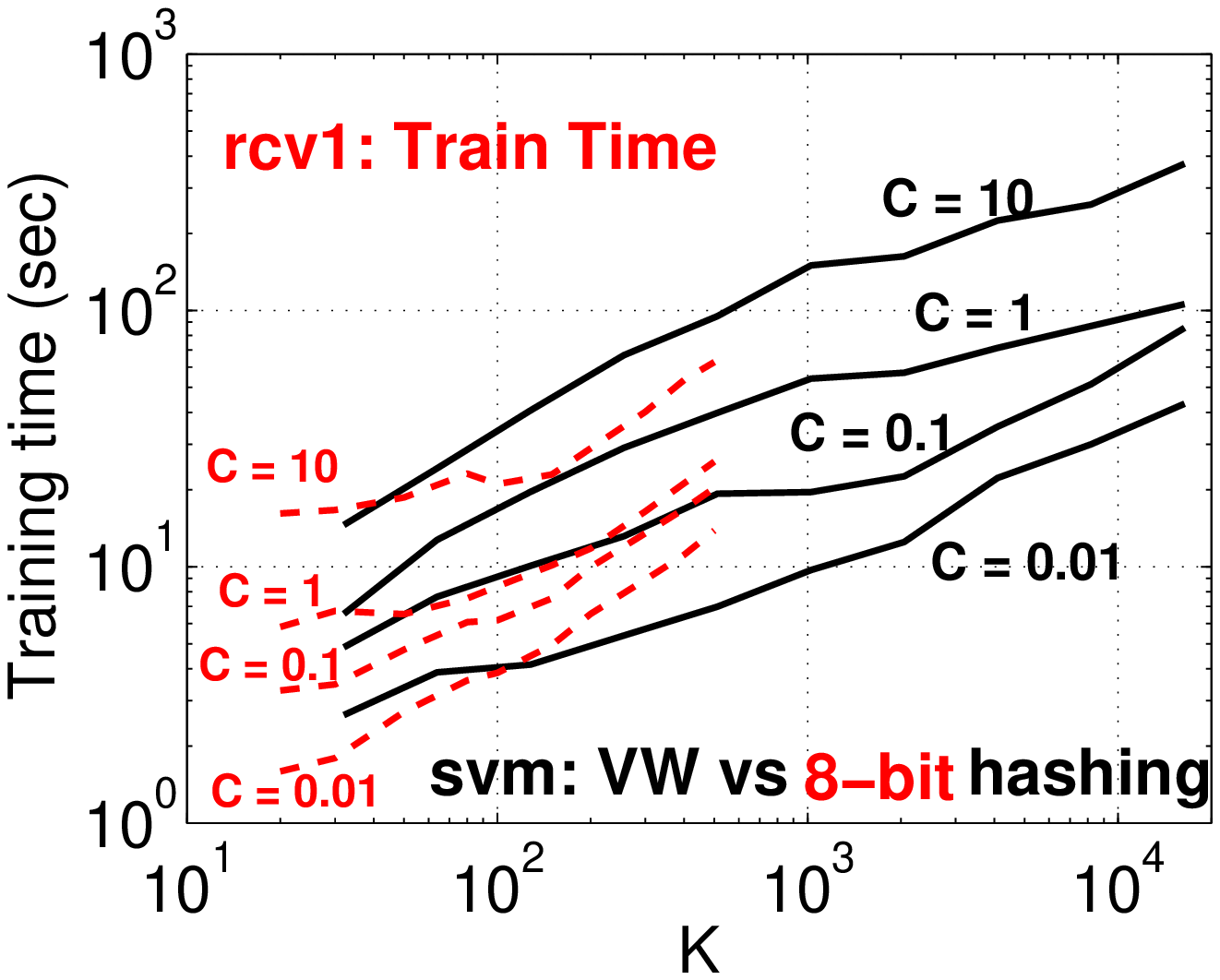}\hspace{0.2in}
\includegraphics[width=1.7in]{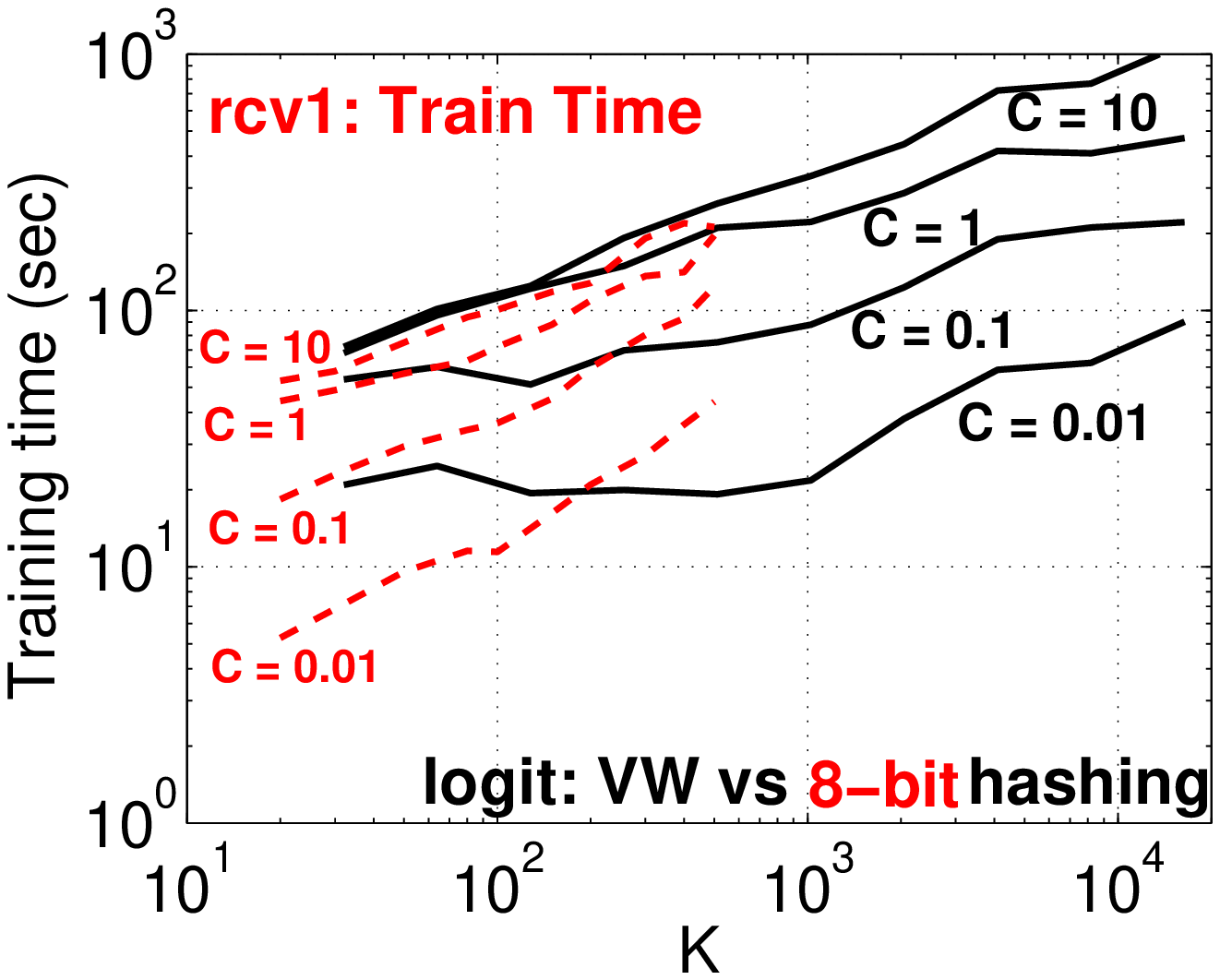}}
\end{center}

\vspace{-0.25in}

\caption{\textbf{Training time for SVM (left) and logistic regression (right) on rcv1} for comparing VW with $8$-bit minwise hashing. }\label{fig_rcv1_train_vw}
\end{figure}

Figure~\ref{fig_rcv1_train_vw} presents the training times for comparing VW with $8$-bit minwise hashing. In this case, we can see that even at the same $k$, 8-bit hashing may have some computational advantages compared to VW. Of course, as it is clear that VW will require a much larger $k$ in order to achieve the same accuracies as 8-bit minwise hashing, we know that the advantage of $b$-bit minwise hashing in terms of training time reduction is also enormous.

\clearpage

Note that, as suggested in~\cite{Report:HashLearning11}, the training time of $b$-bit minwise hashing can be further reduced by applying an additional VW step on top of the data generated by $b$-bit minwise hashing. This is because VW is an excellent tool for achieving {\em compact indexing} when the data dimension is (extremely) much larger than the average number of nonzeros. We conduct the experiments on {\em rcv1} with $b=16$ and notice that this strategy indeed can reduce the training time of $16$-bit minwise hashing by a factor 2 or 3.

\section{Preprocessing Cost}

Minwise hashing has been widely used in (search) industry and $b$-bit minwise hashing requires only very minimal (if any) modifications. Thus, we expect $b$-bit minwise hashing will be adopted in practice. It is also well-understood in practice that we can use (good) hashing functions to very efficiently simulate permutations.

In many real-world scenarios, the preprocessing step is not critical because it requires only one scan of the data, which can be conducted off-line (or on the data-collection stage, or at the same time as n-grams are generated), and it is trivially parallelizable. In fact, because $b$-bit minwise hashing can substantially reduce the memory consumption, it may be now affordable to store considerably more examples in the memory (after $b$-bit hashing) than before,  to avoid (or minimize) disk IOs. Once the hashed data have been generated, they can be used and re-used for many tasks such as supervised  learning, clustering, duplicate detections, near-neighbor search, etc. For example, a learning task may need to re-use the same (hashed) dataset to perform many cross-validations and parameter tuning (e.g., for experimenting with many $C$ values in logistic regression or SVM).   \\

For training truly large-scale datasets, often the data loading time can be dominating~\cite{Proc:Yu_KDD10}. Table~\ref{tab_preprocessing} compares the data loading times with the preprocessing times. For both {\em webspam} and {\em rcv1} datasets, when using a GPU, the preprocessing time for $k=500$ permutations is only a small fraction of the data loading time. Without GPU, the preprocessing time is about 3 or 4 times higher than the data loading time, i.e., they are roughly on the same order of magnitudes. When the training datasets are much larger than 200 GB, we expect that difference between the data loading time and the preprocessing time will be much smaller, even without GPU. We would like to remind that the preprocessing time is only a one-time cost.

\begin{table}[h]
\caption{The data loading and preprocessing (for $k=500$ permutations) times (in seconds).}
\begin{center}{
\begin{tabular}{l l l l }
\hline \hline
Dataset & Data Loading &Preprocessing & Preprocessing with GPU \\\hline
Webspam (24 GB) &$9.7\times10^2$  &$41\times 10^2$ &$0.49\times 10^{2}$  \\
Rcv1 (200 GB) &$1.0\times10^4$  &$3.0\times10^4$  &$0.14\times 10^4$\\
\hline\hline
\end{tabular}
}
\end{center}
\label{tab_preprocessing}
\end{table}

\section{Simulating Permutations Using 2-Universal Hashing}

Conceptually, minwise hashing requires $k$ permutation mappings $\pi_j: \Omega \longrightarrow \Omega$, $j = 1$ to $k$, where $\Omega = \{0, 1, ..., D-1\}$. If we are able to store these $k$ permutation mappings, then the operation is straightforward. For practical industrial applications, however, storing permutations would be infeasible. Instead, permutations are usually simulated by {\em universal hashing}, which only requires  storing very few numbers.

The simplest (and possibly the most popular) approach is to use {\em 2-universal hashing}. That is, we define a series of hashing functions $h_j$ to replace $\pi_j$:
\begin{align}
h_j(t) = \left\{c_{1,j} + c_{2,j}\ t\ \text{  mod  }\ p\right\}\ \text{ mod } \ D, \hspace{0.5in} j = 1, 2, ..., k,
\end{align}
where $p>D$ is a prime number and $c_{1,j}$ is chosen uniformly from $\{0, 1, ..., p-1\}$ and $c_{2,j}$ is chosen uniformly from $\{1, 2, ..., p-1\}$. This way, instead of storing $\pi_j$, we only need to store $2k$ numbers, $c_{1,j}, c_{2,j}$, $j=1$ to $k$. There are several small ``tricks'' for speeding up 2-universal hashing (e.g., avoiding modular arithmetic). An interesting thread  might be {\small\url{http://mybiasedcoin.blogspot.com/2009/12/text-book-algorithms-at-soda-guest-post.html}}

Given a  feature vector (e.g., a document parsed as a list of 1-gram, 2-gram, and 3-grams), for any nonzero location $t$ in the original feature vector, its new location becomes $h_j(t)$; and we walk through all nonzeros locations to find the minimum of the new locations, which will be the $j$th hashed value for that feature vector. Since the generated parameters, $c_{1,j}$ and $c_{2,j}$, are fixed (and stored), this procedure becomes deterministic.\\

\begin{figure}[h!]
\begin{center}
\mbox{
\includegraphics[width=2.5in]{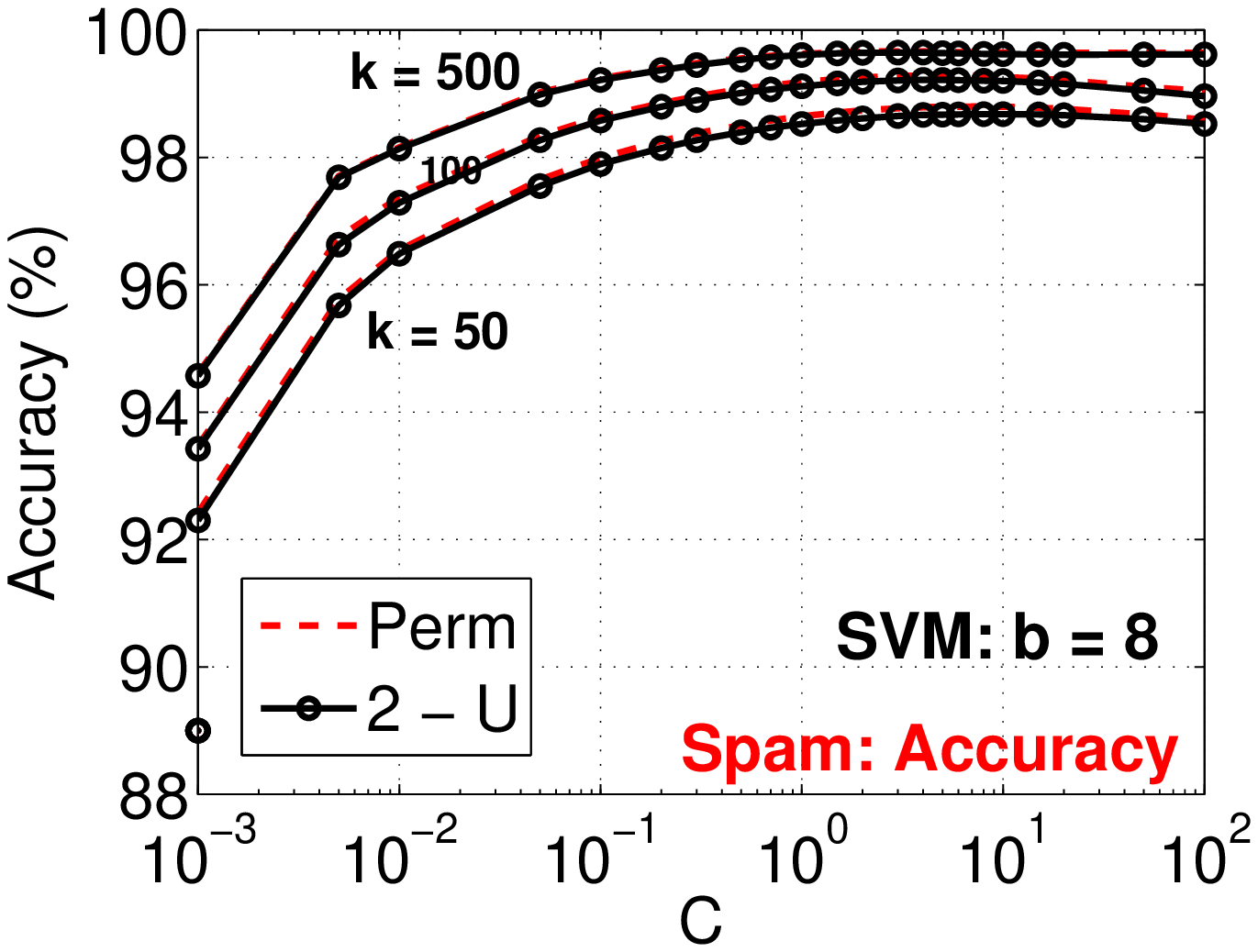}\hspace{0.1in}
\includegraphics[width=2.5in]{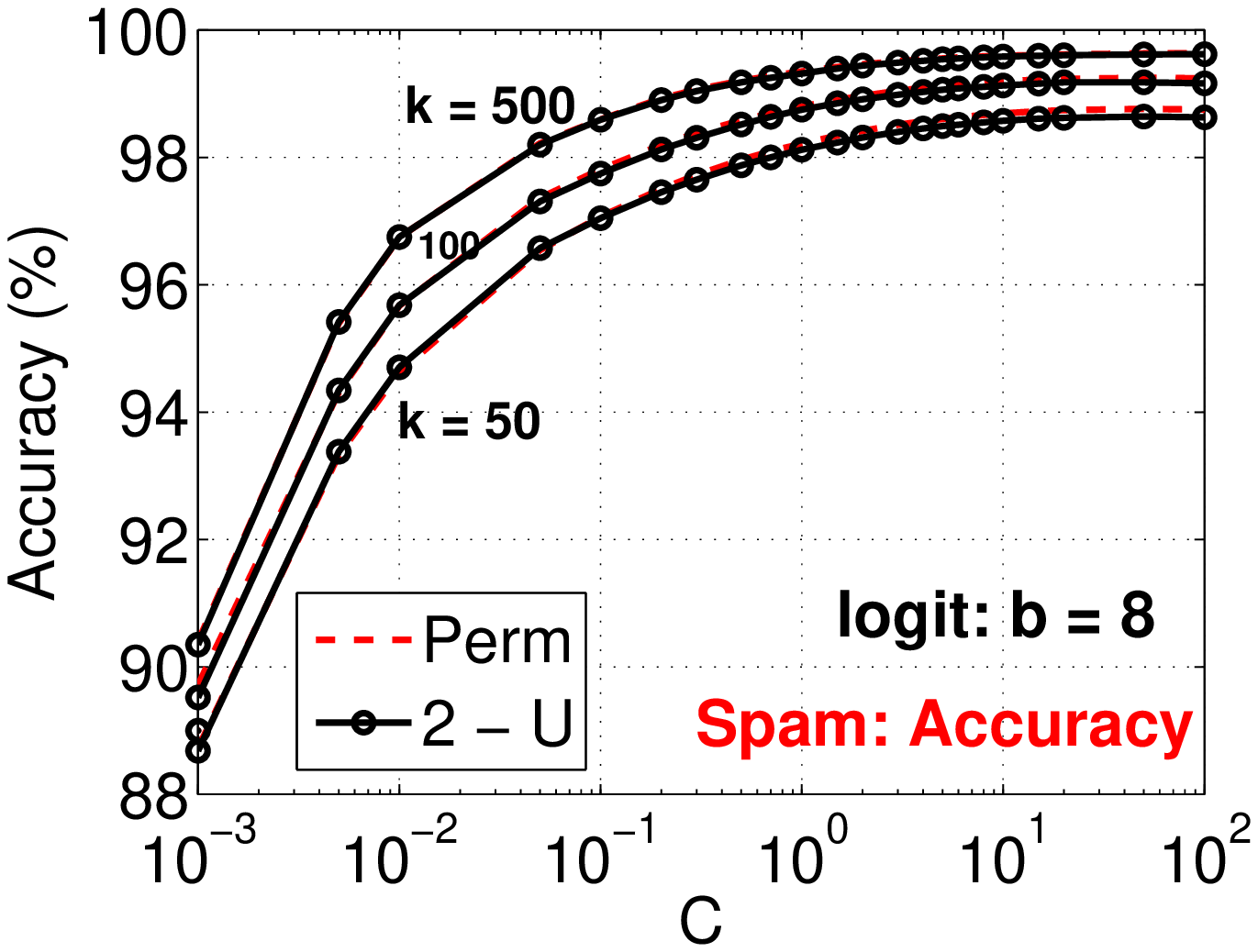}}
\end{center}

\vspace{-0.15in}

\caption{\textbf{Test accuracies on webspam} for comparing permutations (dashed) with 2-universal hashing (solid) (averaged over 50 runs), for both linear SVM (left) and logistic regression (right). We can see that the solid curves essentially overlap the dashed curves, verifying that even the simplest 2-universal hashing method can be very effective.  }\label{fig_spam_acc_2u}
\end{figure}

Our experiments on {\em webspam} can show that even with this simplest hashing method, we can still achieve good performance compared to using perfect random permutations. We can not realistically store $k$ permutations for the {\em rcv1} dataset because its $D = 10^9$. Thus, we only verify the practice of using 2-universal hashing  on the {\em webspam} dataset, as demonstrated in Figure~\ref{fig_spam_acc_2u}.

\section{Conclusion} \label{Conclusion}

It has been a lot of fun to develop $b$-bit minwise hashing and apply it to machine learning for training very large-scale datasets. We hope engineers will find our method applicable to their work. We also hope this work can draw interests from research groups in statistics, theoretical CS, machine learning, or search technology.

{\small

}

\end{document}